\documentclass[runningheads]{llncs}

\usepackage{graphicx}

\usepackage{tikz}
\usepackage{comment}
\usepackage{amsmath,amssymb} 
\usepackage{color}
\usepackage{subcaption}
\usepackage{threeparttable}
\usepackage{array}
\usepackage{booktabs}

\usepackage[accsupp]{axessibility}  

\usepackage[width=122mm,left=12mm,paperwidth=146mm,height=193mm,top=12mm,paperheight=217mm]{geometry}

\usepackage{tikz}
\newcommand\checkmarks[1][]{%
  \tikz[scale=0.4,#1]{\fill(0,.35) -- (.25,0) -- (1,.7) -- (.25,.15) -- cycle;}%
}
\newcommand\crossmark[1][]{%
  \tikz[scale=0.4,#1]{
    \fill(0,0)--(0.1,0) .. controls (0.5,0.4) .. (1,0.7)--(0.9,0.7) ..  controls (0.5,0.5) ..(0,0.1) --cycle;
    \fill(1,0.1)--(0.9,0.1) .. controls (0.5,0.3) .. (0,0.7)--(0.1,0.7) .. controls (0.5,0.4) ..(1,0.2) --cycle;
  }%
}

\usepackage{xcolor,pifont}
\newcommand*\colourcheck[1]{%
  \expandafter\newcommand\csname #1check\endcsname{\textcolor{#1}{\ding{52}}}%
}

\colourcheck{blue}
\colourcheck{green}
\colourcheck{red}
  
\begin{document}

\pagestyle{headings}
\mainmatter
\def\ECCVSubNumber{6192}  

\title{MFIM: Megapixel Facial Identity Manipulation} 
\author{Sangheyon Na\inst{1}}
\institute{Kakao Brain, Seongnam, South Korea\\
\email{orca.na@kakaobrain.com}}

\maketitle
\begin{figure*}
    \begin{center}
    \includegraphics[width=0.95\linewidth]{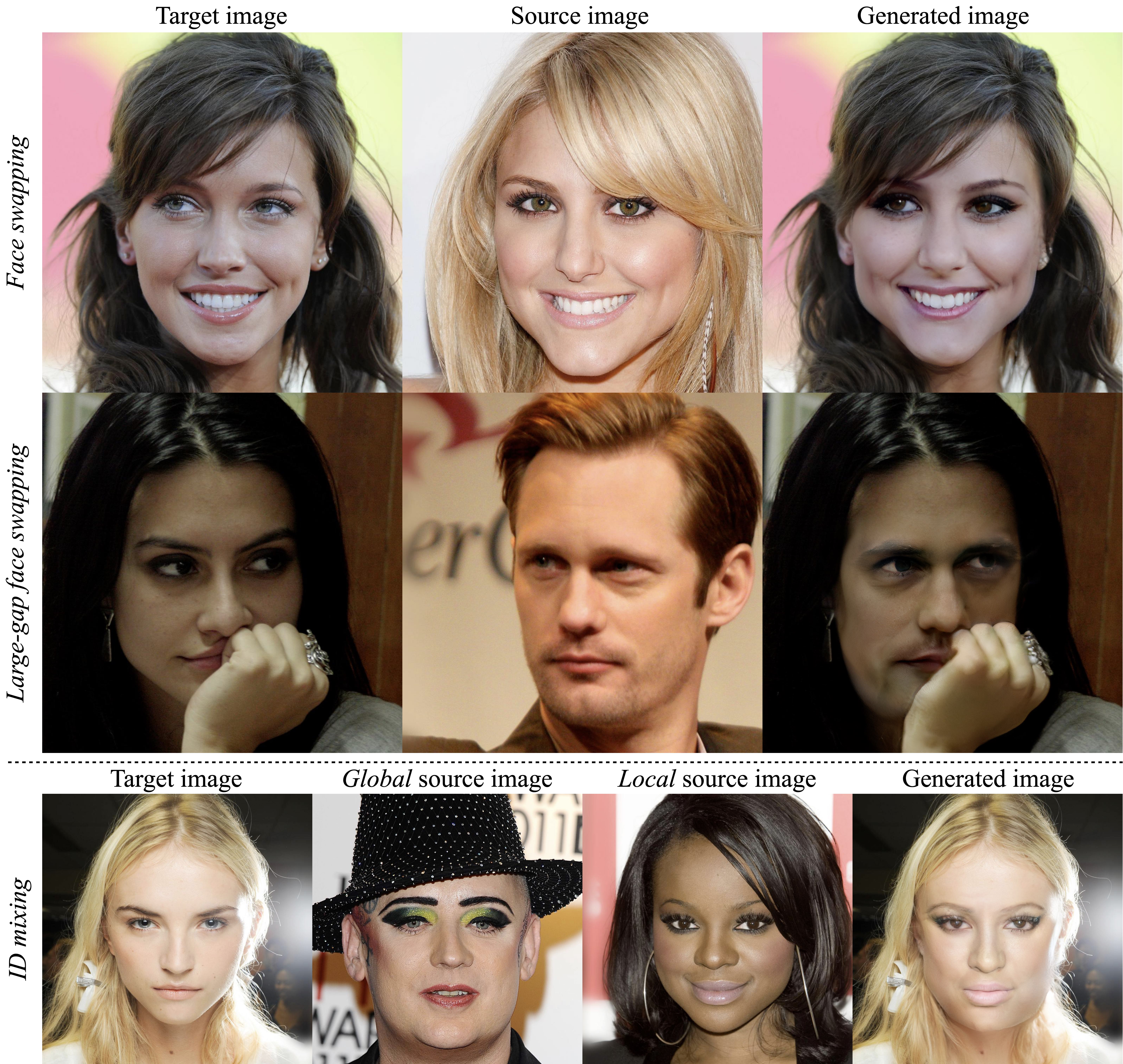}
    \end{center}
    \caption{\textbf{Megapixel facial identity manipulation.} (\textit{Top}) \textbf{Face swapping}. Our model faithfully synthesizes a high-quality megapixel image by blending ID (e.g., eyes and face shape) and ID-irrelevant attributes (e.g., pose and expression) of source and target images, respectively. (\textit{Middle}) \textbf{Face swapping with large gaps} between the source and target images (e.g., gender and age). (\textit{Bottom}) \textbf{ID mixing} using two source images: blending global (e.g., face shape) and local ID attributes (e.g., eyes) of global and local source images, respectively.}
    \label{fig:teaser}
\end{figure*}

\begin{abstract}
Face swapping is a task that changes a facial identity of a given image to that of another person. In this work, we propose a novel face-swapping framework called \textit{Megapixel Facial Identity Manipulation (MFIM)}. The face-swapping model should achieve two goals. First, it should be able to generate a high-quality image. We argue that a model which is proficient in generating a megapixel image can achieve this goal. However, generating a megapixel image is generally difficult without careful model design. Therefore, our model exploits pretrained StyleGAN in the manner of GAN-inversion to effectively generate a megapixel image. Second, it should be able to effectively transform the identity of a given image. Specifically, it should be able to actively transform ID attributes (e.g., face shape and eyes) of a given image into those of another person, while preserving ID-irrelevant attributes (e.g., pose and expression). To achieve this goal, we exploit 3DMM that can capture various facial attributes. Specifically, we explicitly supervise our model to generate a face-swapped image with the desirable attributes using 3DMM. We show that our model achieves state-of-the-art performance through extensive experiments. Furthermore, we propose a new operation called ID mixing, which creates a new identity by semantically mixing the identities of several people. It allows the user to customize the new identity.
\end{abstract}

\section{Introduction}
Face swapping is a task that changes the facial identity of a given image to that of another person. It has now been applied in various applications and services in entertainment~\cite{kemelmacher2016transfiguring}, privacy protection~\cite{mosaddegh2014photorealistic}, and theatrical industry~\cite{naruniec2020high}.

In technical terms, a face-swapping model should be able to generate a high-quality image. At the same time, it should be able to transfer the ID attributes (e.g., face shape and eyes) from the source image to the target image, while preserving the ID-irrelevant attributes (e.g., pose and expression) of the target image as shown in Figure~\ref{fig:teaser}. 
In other words, the face-swapping model has two goals: \lowercase\expandafter{\romannumeral1}) generating high-quality images and \lowercase\expandafter{\romannumeral2}) effective identity transformation. Our model, \textit{Megapixel Facial Identity Manipulation (MFIM)}, is designed to achieve both of these goals.

Firstly, to generate a high-quality image, we propose a face-swapping framework that exploits pretrained StyleGAN~\cite{karras2020analyzing} in the manner of GAN-inversion. Specifically, we design an encoder called facial attribute encoder that effectively extracts ID and ID-irrelevant representations from the source and target images, respectively. These representations are forwarded to the pretrained StyleGAN generator. Then, the generator blends these representations and generates a high-quality megapixel face-swapped image.

Basically, our facial attribute encoder extracts style codes, which is similar to existing StyleGAN-based GAN-inversion encoders~\cite{richardson2021encoding,tov2021designing,alaluf2021restyle}. Specifically, our facial attribute encoder extracts ID and ID-irrelevant style codes from the source and target images, respectively.
Here, one of the important things for faithful face swapping is that the details of the target image such as expression or background should be accurately reconstructed. However, the ID-irrelevant style codes, which do not have spatial dimensions, can fail to preserve the details of the target image. Therefore, our facial attribute encoder extracts not only the style codes, but also the style maps which have spatial dimensions from the target image. The style maps, which take advantages from its spatial dimensions, can complement the ID-irrelevant style codes by propagating additional information about the details of the target image. As a result, our facial attribute encoder, which extracts the style codes and style maps, can effectively capture the ID attributes from the source image and the ID-irrelevant attributes including details from the target image. MegaFS~\cite{zhu2021one}, the previous model that exploits pretrained StyleGAN, suffers from reconstructing the details of target image because it only utilizes the style codes. To solve this problem, they use a segmentation label to take the details from the target image. However, we resolve this drawback by extracting the style maps instead of using the segmentation label.

Secondly, we utilize 3DMM~\cite{DECA:Siggraph2021} which can capture various facial attributes for the effective identity transformation. We especially focus on the transformation of face shape which is one of the important factors in recognizing an identity. However, it is difficult to transform the face shape while preserving the ID-irrelevant attributes of the target image at the same time because these two goals are in conflict with each other~\cite{kim2021smooth}. Specifically, making the generated image have the same face shape with that of the source image enforces the generated image to \textit{differ} a lot from the target image. In contrast, making it preserve the ID-irrelevant attributes of the target image enforces it to be \textit{similar} to the target image. To achieve these two conflicting goals simultaneously, we utilize 3DMM which can accurately and distinctly capture the various facial attributes such as shape, pose, and expression from a given image. In particular, we explicitly supervise our model to generate a face-swapped image with the desirable attributes using 3DMM, i.e., the same face shape with the source image, but the same pose and expression with the target image. The previous models~\cite{deepfakes,li2019faceshifter,chen2020simswap,gao2021information,zhu2021one,kim2021smooth} without such explicit supervision struggle with achieving two conflicting goals simultaneously. In contrast, our model can transform the face shape well, while preserving the ID-irrelevant attributes of the target image. HiFiFace~\cite{wang2021hififace}, the previous model that exploits 3DMM, requires 3DMM not only at the training phase, but even at the inference phase. In contrast, our model does not use 3DMM at the inference phase.

Finally, we propose a new additional task, ID mixing, which means face swapping with a new identity created with multiple source images instead of a single source image. Here, we aim to design a method that allows the user to semantically control the identity creation process. For example, when using two source images, the user can extract the global ID attributes (e.g., face shape) from one source image and the local ID attributes (e.g., eyes) from the other source image, and create the new identity by blending them as shown in Figure~\ref{fig:teaser}. The user can customize the new identity as desired with this operation. Furthermore, this operation does not require any additional training or segmentation label. To the best of our knowledge, we are the first to propose this operation.

In conclusion, the main contributions of this work include the following:

$\bullet$ We propose an improved framework for face swapping by adopting GAN-inversion method with pretrained StyleGAN that takes both style codes and style maps. It allows our model to generate high-quality megapixel images without additional labels in order to preserve the details of the target image.

$\bullet$ We introduce a 3DMM supervision method for the effective identity transformation, especially, the face shape. It allows our model to transform the face shape and preserve the ID-irrelevant attributes at the same time. Moreover, our model does not require 3DMM at the inference phase.

$\bullet$ We propose a new operation, ID mixing, which allows the user to customize the new identity using multiple source images. It does not require any additional training or segmentation label.

\begin{table*}[t]
    \begin{center}
    \caption{\textbf{Comparison of our model (MFIM) with the previous face-swapping models (\greencheck : positive, \crossmark[red, scale=0.8] : negative, \checkmarks[red] : partially positive).} In terms of the 3DMM supervision, HifiFace also exploits the 3DMM supervision, but it requires 3DMM even at the inference phase, while MFIM does not.}
    \label{table:key_differences}
        \begin{tabular}{ccccccc}
            \hline\noalign{\smallskip}
                     & FaceShifter & HifiFace & InfoSwap & MegaFS & SmoothSwap & \textbf{MFIM} \\
            \hline\noalign{\smallskip}
            Megapixel & \crossmark[red, scale=0.8] & \crossmark[red, scale=0.8] & \greencheck & \greencheck & \crossmark[red, scale=0.8] & \greencheck \\
            W/o segmentation labels & \greencheck & \crossmark[red, scale=0.8] & \greencheck & \crossmark[red, scale=0.8] & \greencheck & \greencheck \\
            3DMM supervision & \crossmark[red, scale=0.8] & \checkmarks[red] & \crossmark[red, scale=0.8] & \crossmark[red, scale=0.8] & \crossmark[red, scale=0.8] & \greencheck \\
            ID mixing & \crossmark[red, scale=0.8] & \crossmark[red, scale=0.8] & \crossmark[red, scale=0.8] & \crossmark[red, scale=0.8] & \crossmark[red, scale=0.8] & \greencheck \\
            \hline
        \end{tabular}
    \end{center}
\end{table*}

\section{Related Work}
\label{sec:related_work}
\subsubsection{Face swapping.}
Faceshifter~\cite{li2019faceshifter} proposes a two-stage framework in order to achieve occlusion aware method. Simswap~\cite{chen2020simswap} focuses on designing a framework to transfer an arbitrary identity to the target image. InfoSwap~\cite{gao2021information} proposes explicit supervision based on the IB principle for disentangling identity and identity-irrelevant information from source and target image. MegaFS~\cite{zhu2021one} uses pre-trained StyleGAN~\cite{karras2020analyzing} in order to generate megapixel samples by adopting GAN-inversion method. However, it does not introduce 3DMM supervision and relies on the segmentation labels. HifiFace~\cite{wang2021hififace} utilizes 3DMM for the effective identity transformation. However, HifiFace~\cite{wang2021hififace} requires 3DMM not only in the training phase, but also in the inference phase. On the contrary, our model only takes advantage of 3DMM at training phase and no longer needs it at the inference phase. Most recently, SmoothSwap~\cite{kim2021smooth} proposes a smooth identity embedder to improve learning stability and convergence speed. The key differences between our model and the previous models are given in Table~\ref{table:key_differences}.

\subsubsection{Learning-based GAN-inversion.}
Generative Adversarial Networks (GAN)~\cite{goodfellow2014generative} framework has been actively employed in the various image manipulation applications~\cite{isola2017image,zhu2017unpaired,lee2018diverse,choi2018stargan,choi2020stargan,park2019semantic,bahng2020exploring,na2019miso,cho2019image,yoo2019coloring}. Recently, as remarkable GAN frameworks (e.g., BigGAN~\cite{brock2018large} and StyleGAN~\cite{karras2020analyzing}) have emerged, GAN-inversion~\cite{xia2022gan} is being actively studied. Especially, learning-based GAN-inversion aims to train an extra encoder to find a latent code that can reconstruct a given image using a pretrained generator as a decoder. Then, one can edit the given image by manipulating the latent code. pSp~\cite{richardson2021encoding} and e4e~\cite{tov2021designing} use the pretrained StyleGAN generator as a decoder. However, they have difficulty in accurate reconstruction of the given image. To solve this problem, ReStyle~\cite{alaluf2021restyle} and HFGI~\cite{wang2021high} propose iterative refinement and distortion map, respectively. However, these methods require multiple forward passes. StyleMapGAN~\cite{kim2021exploiting} replaces the style codes of StyleGAN with the style maps. Our model also exploits the style maps, but as additional inputs to the style codes, not as replacements for the style codes to fully utilize the capability of the pretrained StyleGAN generator.

\subsubsection{3DMM.}
A 3D morphable face model (3DMM) produces vector space representations that capture various facial attributes such as shape, expression and pose~\cite{blanz1999morphable,alexander2009digital,cao2013facewarehouse,deng2019accurate,DECA:Siggraph2021}. Although the previous 3DMM methods~\cite{blanz1999morphable,alexander2009digital,cao2013facewarehouse} have limitations in estimating face texture and lighting conditions accurately, recent methods~\cite{deng2019accurate,DECA:Siggraph2021} overcome these limitations. We utilize the state-of-the-art 3DMM~\cite{DECA:Siggraph2021} to effectively capture the various facial attributes and supervise our model.

\section{MFIM: Megapixel Facial Identity Manipulation}

\begin{figure*}[t]
        \begin{subfigure}[b]{0.62\textwidth} 
                \includegraphics[width=\linewidth]{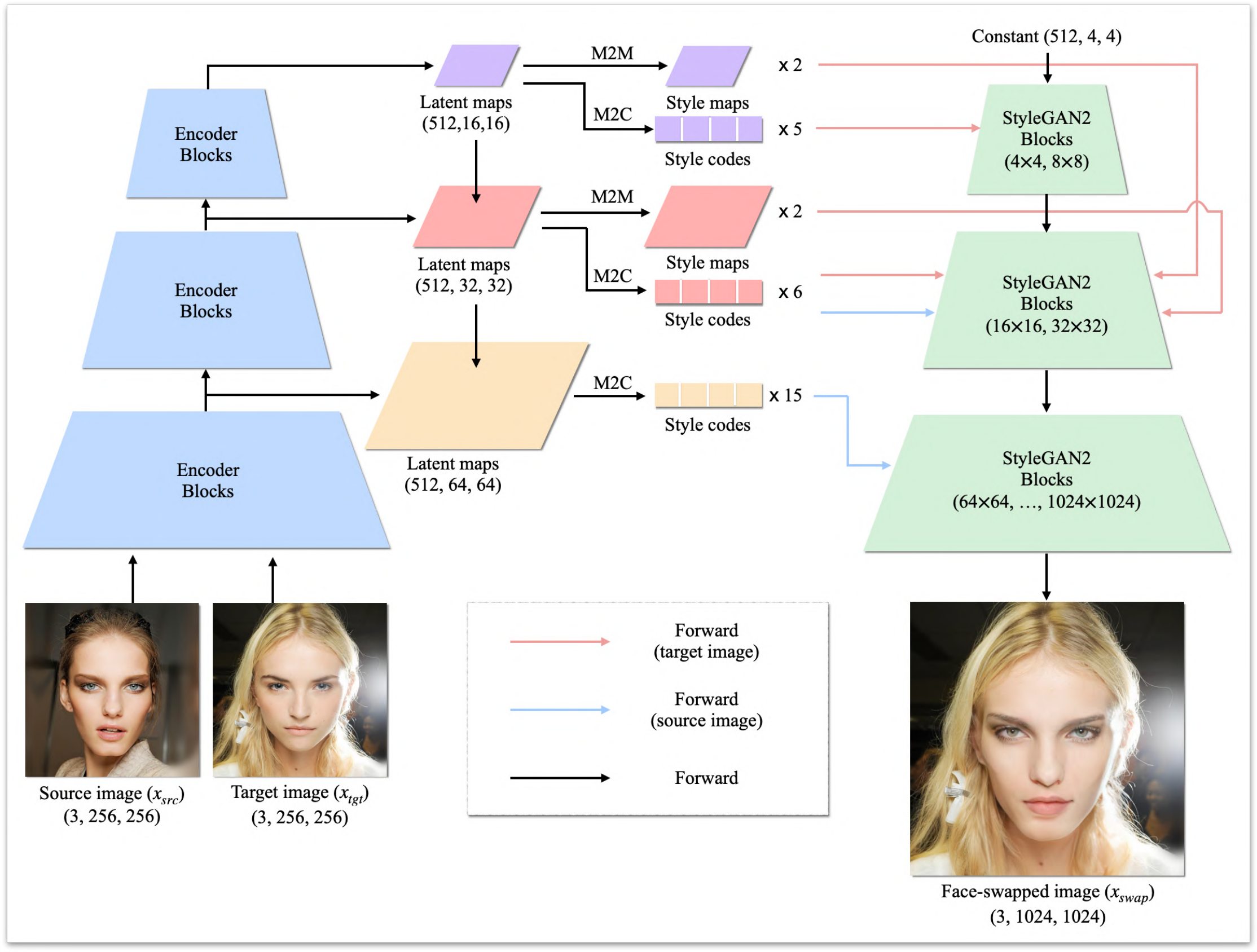}
                \caption{Face swapping}
                \label{subfig:face_swapping}
        \end{subfigure}
        \hfill
        \begin{subfigure}[b]{0.365\textwidth}
                \includegraphics[width=\linewidth]{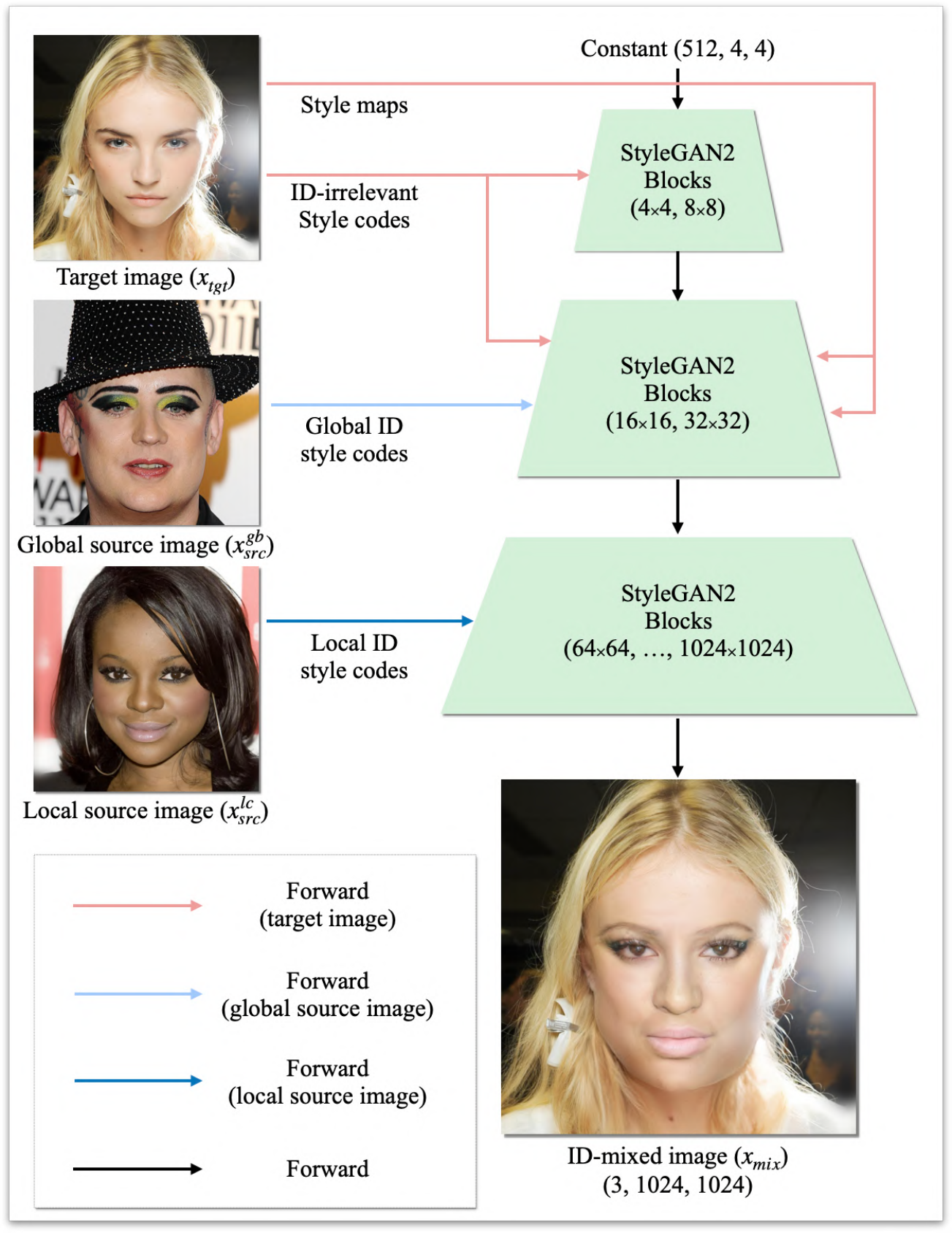}
                \caption{ID mixing}
                \label{subfig:id_mixing}
        \end{subfigure}
        \caption{\textbf{The architecture of MFIM.} Figure~\ref{subfig:face_swapping} shows the process of face swapping. The facial attribute encoder extracts style codes and style maps from source and target images. These are given to the pretrained StyleGAN generator as inputs. Figure~\ref{subfig:id_mixing} shows the process of ID mixing. The ID-style codes are extracted from two source images, instead of a single source image.}
        \label{fig:main_model_figure}
\end{figure*}

Figure~\ref{subfig:face_swapping} shows an overall architecture of our model. Our goal is to capture the ID and ID-irrelevant attributes from the source image, $x_{src} \in \mathbb{R}^{3 \times 256 \times 256}$, and target image, $x_{tgt} \in  \mathbb{R}^{3 \times 256 \times 256}$, respectively, and synthesize a megapixel image, $x_{swap} \in  \mathbb{R}^{3 \times 1024 \times 1024}$, by blending these attributes. Note that $x_{swap}$ should have the same ID attributes with those of $x_{src}$, while the same ID-irrelevant attributes with those of $x_{tgt}$. For example, in Figure~\ref{fig:main_model_figure}, $x_{swap}$ has the same eyes and face shape with $x_{src}$, and the same pose and expression with $x_{tgt}$.

To achieve this goal, we firstly design a facial attribute encoder that encodes $x_{src}$ and $x_{tgt}$ into ID and ID-irrelevant representations, respectively. These representations are forwarded to the pretrained StyleGAN generator (Section~\ref{subsec:facial_attribute_encoder}). Secondly, for the effective identity transformation, especially the face shape, we additionally supervise our model with 3DMM. Note that 3DMM is only used at the training phase and no more used at the inference phase (Section~\ref{subsec:training_objectives}). After training, our model can perform a new operation called \textit{ID mixing} as well as face swapping. Whereas conventional face swapping uses only one source image, ID mixing uses multiple source images to create a new identity. (Section~\ref{subsec:id_mixing}).

\subsection{Facial Attribute Encoder}
\label{subsec:facial_attribute_encoder}
We introduce our facial attribute encoder. As shown in Figure~\ref{subfig:face_swapping}, it first extracts hierarchical latent maps from a given image like pSp encoder~\cite{richardson2021encoding}. Then, map-to-code (M2C) and map-to-map (M2M) blocks produce the style codes and style maps respectively, which are forwarded to the pretrained StyleGAN generator.

\subsubsection{Style code.}
Among the many latent spaces of the pretrained StyelGAN generator (e.g., $\mathcal{Z}$~\cite{karras2019style}, $\mathcal{W}$~\cite{karras2019style}, $\mathcal{W}^+$~\cite{abdal2019image2stylegan}, and $\mathcal{S}$~\cite{wu2021stylespace}), our facial attribute encoder maps a given image to $\mathcal{S}$, so it extracts twenty-six style codes from a given image. The extracted style codes transform the generator feature maps via weight demodulation operation~\cite{karras2020analyzing}. As demonstrated in previous work~\cite{karras2019style}, among the twenty-six style codes, we expect that the style codes corresponding to coarse spatial resolutions (e.g., from $4\times4$ to $16\times16$) synthesize the global aspects of an image (e.g., overall structure and pose). In contrast, the style codes corresponding to fine spatial resolutions (e.g., from $32\times32$ to $1024\times1024$), synthesize the relatively local aspects of an image (e.g., face shape, eyes, nose, and lips). 

Based on this expectation, as shown in Figure~\ref{subfig:face_swapping}, the style codes for the coarse resolutions are extracted from $x_{tgt}$ and encouraged to transfer the global aspects of $x_{tgt}$ such as overall structure and pose. In contrast, the style codes for the fine resolutions are extracted from $x_{src}$ and encouraged to transfer the relatively local aspects of $x_{src}$ such as face shape, eyes, nose, and lips. In this respect, we call the style codes extracted from $x_{tgt}$ and $x_{src}$ ID-irrelevant style codes and ID style codes, respectively. However, it is important to reconstruct the details of the target image (e.g., expression and background), but the ID-irrelevant style codes, which do not have spatial dimensions, lose those details.

\subsubsection{Style map.}
To preserve the details of $x_{tgt}$, our encoder extracts the style maps from $x_{tgt}$ which have the spatial dimensions.
Specifically, the M2M blocks in our encoder produce the style maps with the same size of the incoming latent maps. Then, these style maps are given as noise inputs to the pretrained StyleGAN generator, which are known to generate fine details of the image. 

Note that MegaFS~\cite{zhu2021one} also adopts GAN-inversion method, but it struggles with reconstructing the details of $x_{tgt}$. To solve this problem, it relies on the segmentation label that detects background and mouth to copy those from $x_{tgt}$. In contrast, our model can reconstruct the details of $x_{tgt}$ due to the style maps.

\subsection{Training Objectives}
\label{subsec:training_objectives}

\subsubsection{ID loss.}
To ensure $x_{swap}$ has the same identity with $x_{src}$, we formulate ID loss which calculates cosine similarity between them as
\begin{gather}
    \label{eq:id_loss}
    \mathcal{L}_{id}=1-cos(R(x_{swap}), R(x_{src})),
\end{gather}
where $R$ is the pretrained face recognition model~\cite{deng2019arcface}.

\subsubsection{Reconstruction loss.}
In addition, $x_{swap}$ should be similar to $x_{tgt}$ in most regions except for ID-related regions. To impose this constraint, we define reconstruction loss by adopting pixel-level $L_1$ loss and LPIPS loss~\cite{zhang2018unreasonable} as
\begin{gather}
    \label{eq:recon_loss}
    \mathcal{L}_{recon}= L_1(x_{swap}, x_{tgt}) + LPIPS(x_{swap}, x_{tgt}).
\end{gather}

\subsubsection{Adversarial loss.} To make $x_{swap}$ realistic, we use the non-saturating adversarial loss~\cite{goodfellow2014generative}, $\mathcal{L}_{adv}$, and R1 regularization~\cite{mescheder2018training}, $\mathcal{L}_{R_1}$.

\subsubsection{3DMM supervision.}
We explicitly enforce $x_{swap}$ to have the same face shape with that of $x_{src}$, and same pose and expression with those of $x_{tgt}$. For these constraints, we formulate the following losses using 3DMM~\cite{DECA:Siggraph2021}:
\begin{gather}
    \label{eq:3dmm_loss}
        \mathcal{L}_{shape}= ||s_{swap} - s_{src} ||_2, \\
        \mathcal{L}_{pose}= ||p_{swap} - p_{tgt} ||_2, \\
        \mathcal{L}_{exp}= ||e_{swap} - e_{tgt} ||_2,
\end{gather}
where $s$, $p$, and $e$ are the shape, pose, and expression parameters extracted from a given image by 3DMM~\cite{DECA:Siggraph2021} encoder, respectively, with a subscript that denotes the image from which the parameter is extracted (e.g., $s_{swap}$ is the shape parameter extracted from $x_{swap}$). $\mathcal{L}_{shape}$ encourages $x_{swap}$ to have the same face shape with that of $x_{src}$. On the other hand, $\mathcal{L}_{pose}$ and $\mathcal{L}_{exp}$ encourage $x_{swap}$ to have the same pose and expression with those of $x_{tgt}$, respectively.

Note that HifiFace~\cite{wang2021hififace} also utilizes 3DMM, but it requires 3DMM even at the inference phase. This is because HiFiFace takes 3DMM parameters as inputs to generate a face-swapped image. In contrast, our model does not take 3DMM parameters as inputs to generate a face-swapped image, so 3DMM is no more used at the inference phase. Furthermore, in terms of loss function, HifiFace formulates the landmark-based loss, but we formulate the parameter-based losses. We compare these methods in the supplementary material.

\begin{figure*}[t]
    \begin{center}
    \includegraphics[width=1\linewidth]{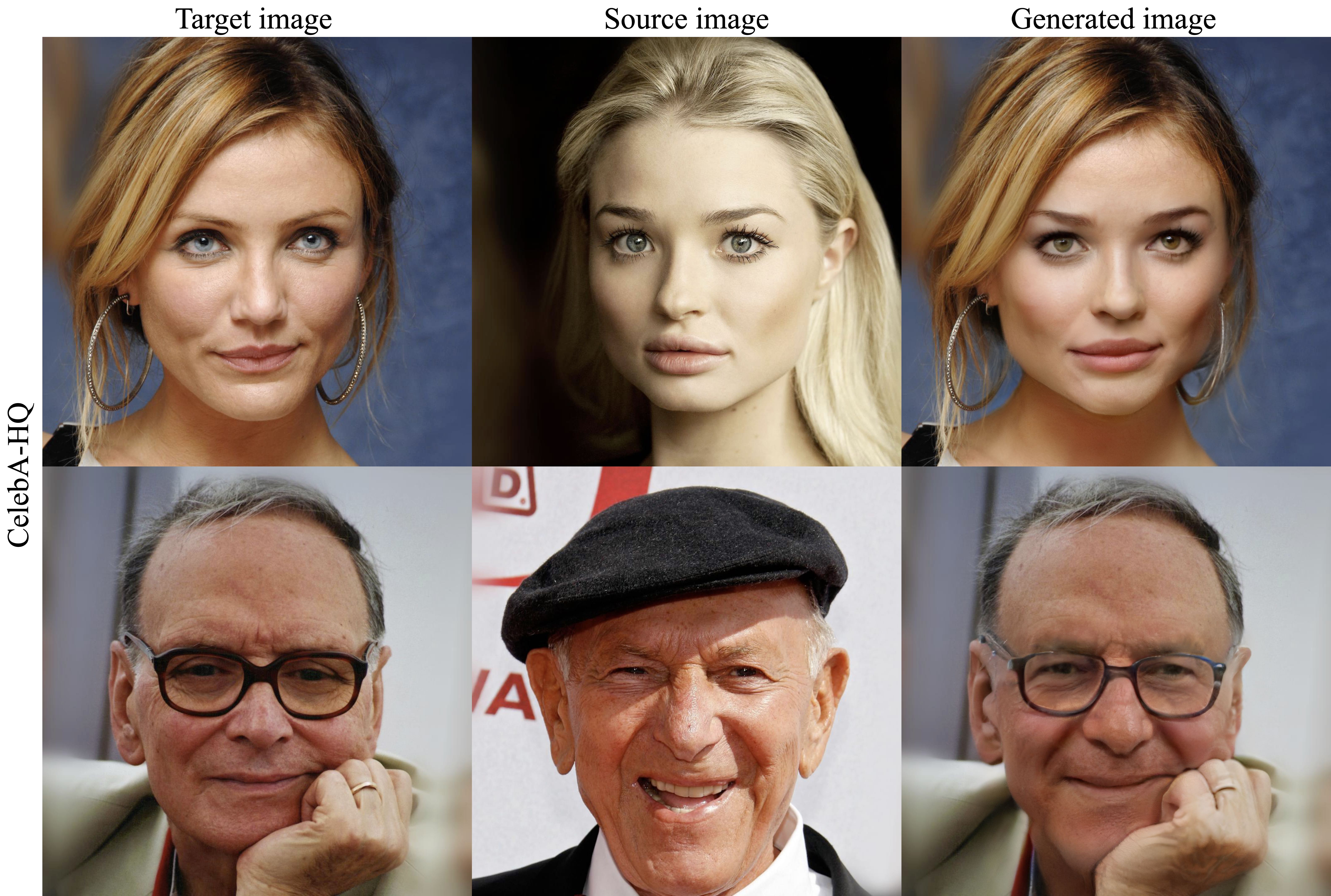}
    \end{center}
    \caption{\textbf{Qualitative results on CelebA-HQ.} The generated images have the same ID attributes (e.g., face shape and eyes) with the source images, but the same ID-irrelevant attributes (e.g., pose and expression) with the taget images.}
    \label{fig:qualitative_results}
\end{figure*}

\subsubsection{Full objective.}
Finally, we formulate the full loss as
\begin{gather}
    \label{eq:final_loss}
        \mathcal{L}=\lambda_{id}\mathcal{L}_{id} + \lambda_{recon}\mathcal{L}_{recon} + \lambda_{adv}\mathcal{L}_{adv} + \lambda_{R_1}\mathcal{L}_{R_1} \nonumber \\
        + \lambda_{shape}\mathcal{L}_{shape} + \lambda_{pose}\mathcal{L}_{pose} + \lambda_{exp}\mathcal{L}_{exp}.
\end{gather}

\begin{figure*}[t]
    \begin{center}
    \includegraphics[width=1\linewidth]{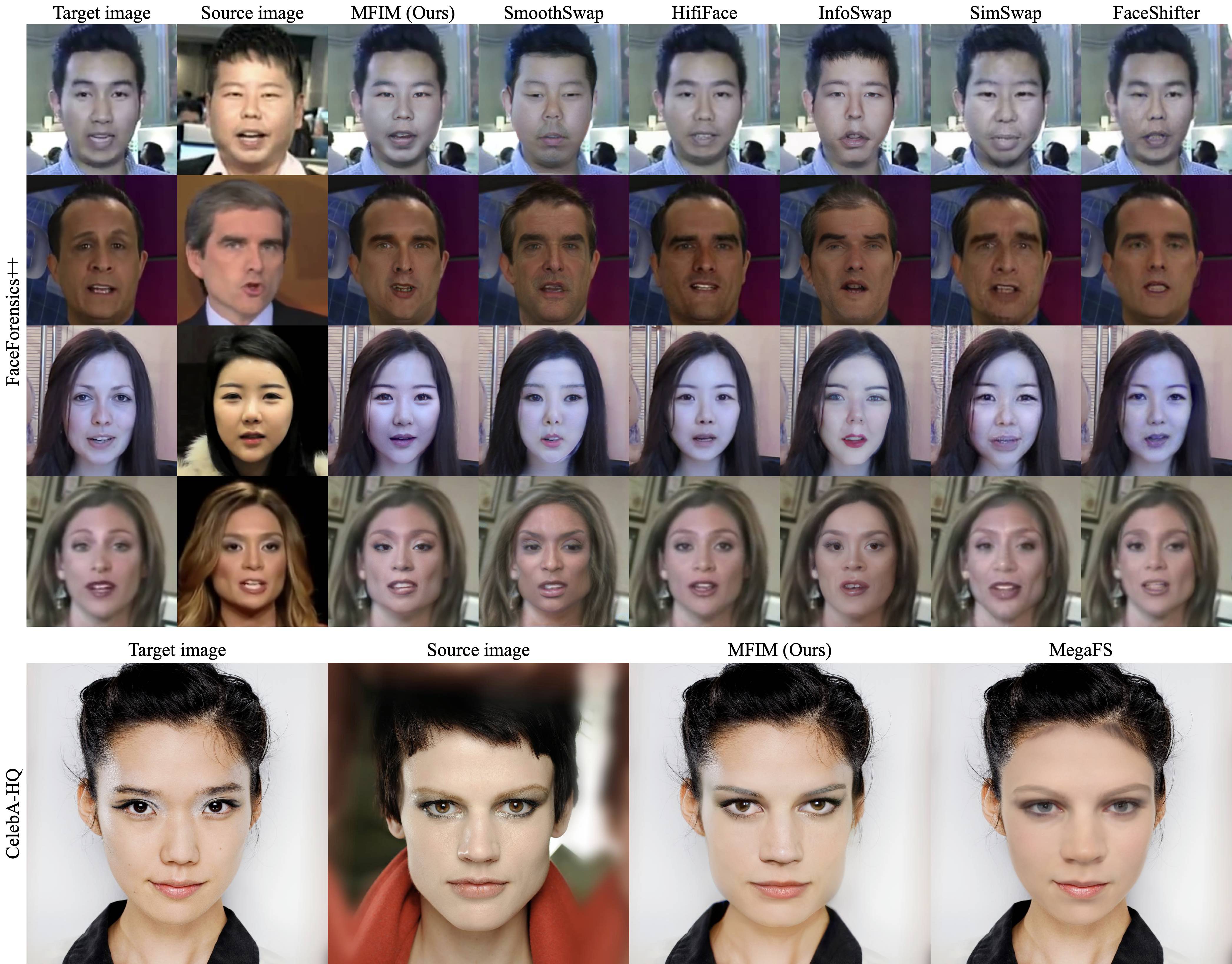}
    \end{center}
    \caption{\textbf{Qualitative comparison.} See Section~\ref{subsec:comparison_with_the_baselines} for the discussion.}
    \label{fig:qualitative_comparison}
\end{figure*}

\subsection{ID Mixing}
\label{subsec:id_mixing}
Our model can create a new identity by mixing multiple identities. We call this operation \textit{ID mixing}. In order to allow the user to semantically control the identity creation process, we design a method to extract the ID style codes from multiple source images and then mix them like style mixing~\cite{karras2019style}. Here, we describe ID mixing using two source images, but it can be generalized to use multiple source images more than two. Specifically, when using two source images, the user can take \textit{global} ID attributes (e.g., face shape) from one source image and \textit{local} ID attributes (e.g., eyes) from the other source image and mix them to synthesize an ID-mixed image, $x_{mix}$.

Figure~\ref{subfig:id_mixing} describes this process. The ID-irrelevant style codes and style maps are extracted from $x_{tgt}$ (red arrow in Figure~\ref{subfig:id_mixing}). However, the ID style codes are extracted from two source images, global and local source images. We denote them as $x_{src}^{gb}$ and $x_{src}^{lc}$, respectively, and the style codes extracted from them are called global (light blue arrow in Figure~\ref{subfig:id_mixing}) and local ID style codes (dark blue arrow in Figure~\ref{subfig:id_mixing}), respectively. These ID style codes transform the specific generator feature maps. In particular, the global ID style codes transform the ones with coarse spatial resolution (e.g., $32 \times 32$), while the local ID style codes are for the ones with fine spatial resolutions (e.g., from $64 \times 64$ to $1024 \times 1024$). In this manner, the global ID style codes transfer the global ID attributes (e.g., face shape) of $x_{src}^{gb}$, while the local ID style codes transfer the local ID attributes (e.g., eyes) of $x_{src}^{lc}$ due to the property of style localization~\cite{karras2019style}. 

MegaFS~\cite{zhu2021one} which exploits pretrained StyleGAN also has the potential to perform ID mixing. However, MegaFS struggles with transforming the face shape~(Section~\ref{subsec:comparison_with_the_baselines}), so it is difficult to effectively perform ID mixing.

\begin{table*}[t]
    \begin{center}
    \caption{\textbf{Quantitative comparison on FaceForensics++.} See Section~\ref{subsec:evaluation_metrics} for the description of each metric, and Section~\ref{subsec:comparison_with_the_baselines} for the discussion.}
    \label{table:quantitative_comparison_ff}
        \setlength{\tabcolsep}{0.8em}
        \scalebox{1}{
            \begin{tabular}{c|ccccc}
                \hline\noalign{\smallskip}
                         & Identity $\downarrow$ & Shape $\downarrow$ & Expression $\downarrow$ & Pose $\downarrow$ & Pose-HN $\downarrow$ \\
                \hline\noalign{\smallskip}
                Deepfakes & 120.907 & 0.639 & 0.802 & 0.188 & 4.588 \\
                FaceShifter & 110.875 & 0.658 & 0.653 & 0.177 & \textbf{3.175} \\
                SimSwap & 99.736 & 0.662 & 0.664 & 0.178 & 3.749 \\
                HifiFace & 106.655 & 0.616 & 0.702 & 0.177 & 3.370 \\
                InfoSwap & 104.456 & 0.664 & 0.698 & 0.179 & 4.043 \\
                MegaFS & 110.897 & 0.701 & 0.678 & 0.182 & 5.456 \\
                SmoothSwap & 101.678 & 0.565 & 0.722 & 0.186 & 4.498 \\
                MFIM (ours) & \textbf{87.030} & \textbf{0.553} & \textbf{0.646} & \textbf{0.175} & 3.694 \\
                \hline
            \end{tabular}
        }
    \end{center}
\end{table*}

\begin{table*}[t]
    \begin{center}
    \caption{\textbf{Quantitative comparison on CelebA-HQ.} See Section~\ref{subsec:evaluation_metrics} for the description of each metric, and Section~\ref{subsec:comparison_with_the_baselines} for the discussion.}
    \label{table:quantitative_comparison_celeba_hq}
        \setlength{\tabcolsep}{0.5em}
        \scalebox{1}{
            \begin{tabular}{c|cccccc}
                \hline\noalign{\smallskip}
                         & Identity $\downarrow$ & Shape $\downarrow$ & Expression $\downarrow$ & Pose $\downarrow$ & Pose-HN $\downarrow$ & FID $\downarrow$ \\
                \hline\noalign{\smallskip}
                MegaFS & 108.571 & 0.906 & 0.438 & 0.071 & 4.880 & 14.446 \\
                MFIM (ours) & \textbf{91.469} & \textbf{0.782} & \textbf{0.400} & \textbf{0.057} & \textbf{4.095} & \textbf{4.946} \\
                \hline
            \end{tabular}
        }
    \end{center}
\end{table*}


\section{Experiments}
We present our experimental settings and results to demonstrate the effectiveness of our model. Implementation details are in the supplementary material.

\subsection{Experimental Settings}
\subsubsection{Baselines.}
We compare our model with Deepfakes~\cite{deepfakes}, FaceShifter~\cite{li2019faceshifter}, SimSwap~\cite{chen2020simswap}, HifiFace~\cite{wang2021hififace}, InfoSwap~\cite{gao2021information}, MegaFs~\cite{zhu2021one}, and SmoothSwap~\cite{kim2021smooth}.

\subsubsection{Datasets.}
We use FFHQ~\cite{karras2019style} for training, and FaceForensics++~\cite{rossler2019faceforensics++} and CelebA-HQ~\cite{karras2017progressive} for evaluation. We do not extend the training dataset by combining multiple datasets, while some of the previous models~\cite{li2019faceshifter,wang2021hififace,zhu2021one,gao2021information} do.

\subsubsection{Evaluation metrics.}
\label{subsec:evaluation_metrics}
We evaluate our model and the baselines with respect to identity, shape, expression, and pose following SmoothSwap~\cite{kim2021smooth}. In the case of ID and shape, the closer $x_{swap}$ and $x_{src}$ are, the better, and for the expression and pose, the closer $x_{swap}$ and $x_{tgt}$ are, the better. To measure the identity, we use $L_2$ distance in the feature space of the face recognition model~\cite{cao2018vggface2}. On the other hand, to measure the shape, expression, and pose, we use $L_2$ distance in the parameter space of 3DMM~\cite{sanyal2019learning} for each attribute. For the pose, $L_2$ distance in the feature space of a pose estimation model~\cite{Ruiz_2018_CVPR_Workshops} is additionally used, and this score is denoted as pose-HN. All of these metrics are the lower the better. 

\begin{figure*}[t]
    \begin{center}
    \includegraphics[width=1\linewidth]{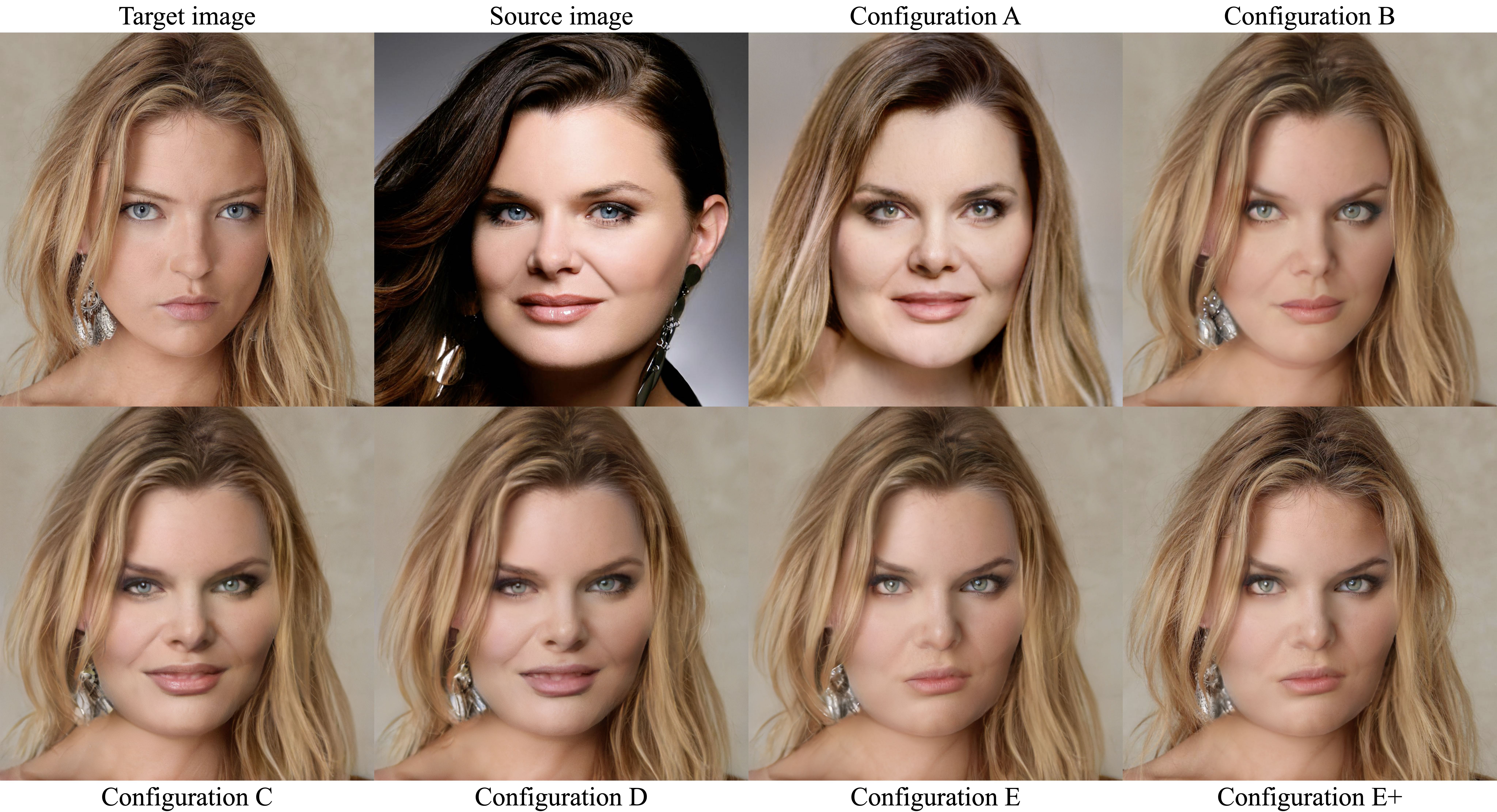}
    \end{center}
    \caption{\textbf{Ablation study of MFIM.} See Section~\ref{subsec:analysis_of_mfim} for the discussion.}
    \label{fig:ablation_study}
\end{figure*}

\subsection{Comparison with the Baselines}
\label{subsec:comparison_with_the_baselines}
The generated images of our model can be seen in Figure~\ref{fig:qualitative_results}. The qualitative and quantitative comparisons between our model and the baselines are presented in Figure~\ref{fig:qualitative_comparison} and Tables~\ref{table:quantitative_comparison_ff} and \ref{table:quantitative_comparison_celeba_hq}, respectively. We first compare our model to the baselines on FaceForensics++~\cite{rossler2019faceforensics++}, following the evaluation protocol of SmoothSwap~\cite{kim2021smooth}. As shown in Table~\ref{table:quantitative_comparison_ff}, our model is superior to the baselines in all metrics except for pose-HN. It is noteworthy that our model outperforms the baselines for the shape, expression, and pose at the same time, whereas the existing baselines do not perform well for all those three metrics at the same time. For example, among the baslines, SmoothSwap~\cite{kim2021smooth} and HifiFace~\cite{wang2021hififace} achieve good scores in the shape, but the expression scores of these baselines are not as good. On the other hand, FaceShifter~\cite{li2019faceshifter} and SimSwap~\cite{chen2020simswap} achieve good scores in the expression and pose, but the shape scores of these baselines are not as good. However, our model accomplishes the state-of-the-art performance for the shape, expression, and pose metric at the same time.

In addition, we compare our model to the previous megapixel model, MegaFS~\cite{zhu2021one}, on CelebA-HQ. We generate 300,000 images following MegaFS~\cite{zhu2021one}. Then, each model is evaluated with the same metrics used in the evaluation on FaceForenscis++. For FID, we use CelebA-HQ for the real distribution following MegaFS~\cite{zhu2021one}. As shown in Table~\ref{table:quantitative_comparison_celeba_hq}, our model outperforms MegaFS~\cite{zhu2021one} in the all metrics.

\begin{table*}[t]
    \begin{center}
    \caption{\textbf{Ablation study of MFIM.} See Section~\ref{subsec:evaluation_metrics} for the description of each metric, and Section~\ref{subsec:analysis_of_mfim} for the discussion.}
    \label{table:ablation_study}
        \setlength{\tabcolsep}{0.7em}
        \scalebox{1}{
            \begin{tabular}{l|ccccc}
                \hline\noalign{\smallskip}
                $~~~~$Configuration & Identity $\downarrow$ & Shape $\downarrow$ & Expression $\downarrow$ & Pose $\downarrow$ & Pose-HN $\downarrow$ \\
                \hline\noalign{\smallskip}
                \textrm{A}. Baseline MFIM & 70.160 & 0.383 & 1.116 & 0.145 & 7.899 \\
                \textrm{B}. $+$ style maps & 91.430 & 0.823 & 0.398 & 0.051 & 3.795 \\
                \textrm{C}. $+~\mathcal{L}_{shape}$ & 86.476 & 0.635 & 0.864 & 0.085 & 5.091 \\
                \textrm{D}. $+~\mathcal{L}_{pose}$ & 86.777 & 0.634 & 0.860 & 0.078 & 4.797 \\
                \textrm{E}. $+~\mathcal{L}_{exp}$ & 91.469 & 0.782 & 0.400 & 0.057 & 4.095 \\
                \hline
            \end{tabular}
            }
    \end{center}
\end{table*}

\subsection{Ablation Study of MFIM}
\label{subsec:analysis_of_mfim}

We conduct an ablation study on CelebA-HQ to demonstrate the effectiveness of each component of our model following the evaluation protocol of the comparative experiment on CelebA-HQ~(Section~\ref{subsec:comparison_with_the_baselines}). The qualitative and quantitative results are presented in Figure~\ref{fig:ablation_study} and Table~\ref{table:ablation_study}, respectively.

The configuration (A) is trained by using only the ID-irrelevant and ID style codes. The style maps and 3DMM supervision are not used in this configuration. In Figure~\ref{fig:ablation_study}, the configuration (A) generates an image that has the overall structure and pose of $x_{tgt}$, but has the identity of $x_{src}$ (e.g., eyes and face shape). This is because the ID-irrelevant style codes transform the generator feature maps with the coarser spatial resolutions (from $4\times4$ to $16\times16$) than the ID style codes (from $32\times32$ to $1024\times1024$), so the ID-irrelevant style codes synthesize more global aspects than the ID style codes do. However, the configuration (A) fails to reconstruct the details of $x_{tgt}$ (e.g., expression, hair style, and background). This is because the ID-irrelevant style codes, which do not have the spatial dimensions, lose the details of $x_{tgt}$.

To solve this problem, we construct the configuration (B) by adding the style maps to the configuration (A). In Figure~\ref{fig:ablation_study}, the configuration (B) reconstructs the details of $x_{tgt}$ better than configuration (A). It is also supported by the improvement of the expression score in Table~\ref{table:ablation_study}. These results show that the style maps, which have the spatial dimensions, can preserve the details of $x_{tgt}$. However, the generated image by configuration (B) does not have the same face shape with that of $x_{src}$, but with that of $x_{tgt}$.

Therefore, for the more effective identity transformation, we improve our model by adding the 3DMM supervision to the configuration (B). First, we construct the configuration (C) by adding $\mathcal{L}_{shape}$ to the configuration (B). As a result, the generated image by the configuration (C) has the same face shape with that of $x_{src}$ rather than that of $x_{tgt}$. It leads to the improvement of the shape score in Table~\ref{table:ablation_study}. However, the expression and pose scores are degraded. This result is consistent with Figure~\ref{fig:ablation_study} in that the generated image of configuration (C) has the same expression with $x_{src}$, not $x_{tgt}$, which is undesirable. We assume that this is because the expression and pose of $x_{src}$ are leaked somewhat while the face shape of $x_{src}$ is actively transferred by $L_{shape}$. It means that the ID and ID-irrelevant representations of MFIM are not perfectly disentangled. Improving our model to solve this problem can be future work.

In order to restore the pose and expression scores, we first construct the configuration (D) by adding $\mathcal{L}_{pose}$ to the configuration (C), and then construct the configuration (E) by adding $\mathcal{L}_{exp}$ to the configuration (D). As a result, as shown in Table~\ref{table:ablation_study}, the pose and expression scores are restored to the similar scores to the configuration (B). Finally, the generated image by the configuration (E) in Figure~\ref{fig:ablation_study} has the same face shape with that of $x_{src}$, while the same pose and expression with that of $x_{tgt}$. 

Although the configuration (E) can faithfully reconstruct the details of $x_{tgt}$ such as background and hair style, we can further improve our model to reconstruct the high-frequency details by adding ROI only synthesis to the configuration (E) at the inference phase. This configuration is denoted as (E+). It allows our model to generate only the face region, but it does not require any segmentation label. More details on this are in the supplementary material. In Figure~\ref{fig:ablation_study}, the configuration (E+) reconstructs the high-frequency details on hair. We use the configuration (E) for all the quantitative results, and the configuration (E+) for all the qualitative results.

\begin{figure*}[t]
    \begin{center}
    \includegraphics[width=1\linewidth]{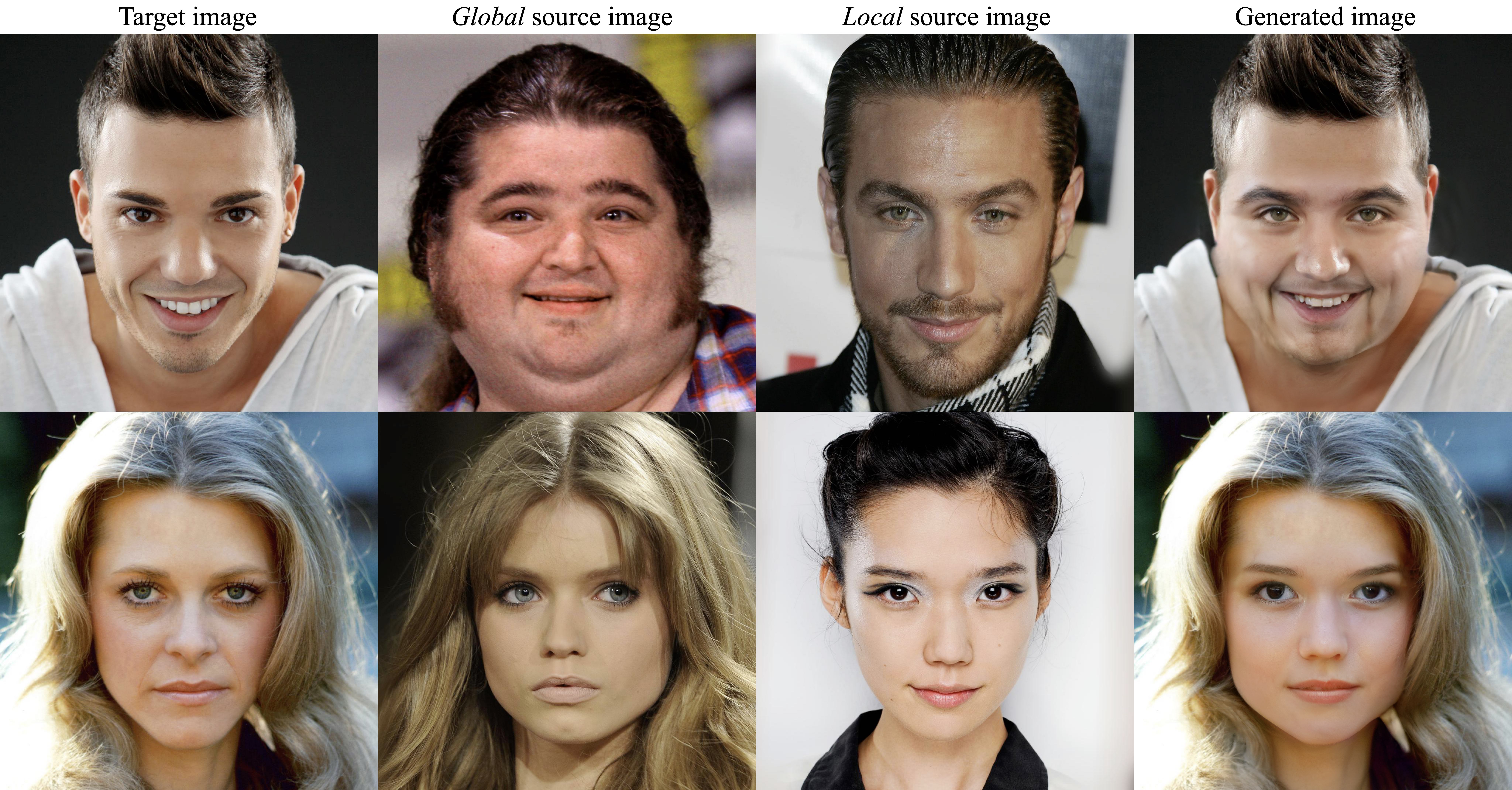}
    \end{center}
    \caption{\textbf{ID mixing.} Our model can create a new identity by blending the global (e.g., face shape) and local (e.g., eyes) ID attributes captured from the global and local source images, respectively.}
    \label{fig:id_mixing}
\end{figure*}

\subsection{ID Mixing}
\label{subsec:id_mixing_analysis}

Figure~\ref{fig:id_mixing} shows the qualitative results of ID mixing using our model. In Figure~\ref{fig:id_mixing}, $x_{mix}$ has the new identity with the global ID attributes (e.g., face shape) of $x_{src}^{gb}$, but the local ID attributes (e.g., eyes) of $x_{src}^{lc}$. This property of ID mixing allows the user to semantically control the ID creation process. We also compare our model with MegaFS~\cite{zhu2021one} in terms of ID mixing in the supplementary material.

We quantitatively analyze the properties of ID mixing on CelebA-HQ. We prepare 30,000 triplets by randomly assigning one global source image and one local source image to each target image. Then, we define Relative Identity ($R\text{-}ID$) distance and Relative Shape ($R\text{-}Shape$) distance following SmoothSwap~\cite{kim2021smooth}. For example, $R\text{-}ID(gb)$ is defined as $R\text{-}ID(gb) = \frac{D_{ID}(x_{mix},~x_{src}^{gb})}{D_{ID}(x_{mix},~x_{src}^{gb}) + D_{ID}(x_{mix},~x_{src}^{lc})}$ where $D_{ID}$ means $L_2$ distance on the feature space of the face recognition model~\cite{cao2018vggface2}. This measures how similar the overall identity of $x_{mix}$ is to that of $x_{src}^{gb}$ compared to $x_{src}^{lc}$. $R\text{-}ID(lc)$ is defined similarly, so $R\text{-}ID(gb) + R\text{-}ID(lc) = 1$. In addition, $R\text{-}Shape(gb)$ and $R\text{-}Shape(lc)$ are defined in the same manner with $R\text{-}ID(gb)$ and $R\text{-}ID(lc)$, respectively, but they are based on the 3DMM~\cite{sanyal2019learning} shape parameter distance to measure the similarity of face shape. 

In Table~\ref{table:id_mixing}, the two rows denoted by \textit{local} and \textit{global} show the results of conventional face swapping, not ID mixing, which uses a single source image. In particular, the row denoted by local is the result of conventional face swapping using only $x_{src}^{lc}$ as the source image without using $x_{src}^{gb}$. For this reason, $R\text{-}ID(lc)$ and $R\text{-}Shape(lc)$ are smaller than $R\text{-}ID(gb)$ and $R\text{-}Shape(gb)$, respectively, which means that the generated image has the same overall identity and face shape as $x_{src}^{lc}$, not $x_{src}^{gb}$. Similarly, the row denoted by global shows that $x_{mix}$ has the same overall identity and face shape as $x_{src}^{gb}$, not $x_{src}^{lc}$. 

On the other hand, the row denoted by ID mixing shows the results of ID mixing, which uses both the $x_{src}^{gb}$ and $x_{src}^{lc}$ as described in Section~\ref{subsec:id_mixing}. In contrast to when only one of $x_{src}^{lc}$ or $x_{src}^{gb}$ is used, $R\text{-}ID(gb)$ is similar to that of $R\text{-}ID(lc)$. It means that the overall identity of $x_{mix}$ by ID mixing is like a new identity, a mixed identity of $x_{src}^{lc}$ and $x_{src}^{gb}$. Furthermore, $R\text{-}Shape(gb)$ has a smaller value than $R\text{-}Shape(lc)$. It means that the face shape of the generated image is more similar to that of $x_{src}^{gb}$ than that of $x_{src}^{lc}$, which is consistent with Figure~\ref{fig:id_mixing}.

\begin{table*}[t]
    \begin{center}
    \caption{\textbf{Quantitative analysis of ID mixing.} See Section~\ref{subsec:id_mixing_analysis} for the description of each metric and discussion.}
    \label{table:id_mixing}
        \setlength{\tabcolsep}{0.7em}
        \scalebox{1}{
            \begin{tabular}{c|cccc}
                \hline\noalign{\smallskip}
                & \multicolumn{2}{c}{Overall identity} & \multicolumn{2}{c}{Face shape} \\
                \hline\noalign{\smallskip}
                & R-ID (gb) & R-ID (lc) & R-Shape (gb) & R-Shape (lc) \\ 
                \hline\noalign{\smallskip}
                Local & 0.602 & 0.398 & 0.609 & 0.391 \\ 
                ID mixing & 0.515 & 0.485 & 0.466 & 0.534 \\ 
                Global & 0.399 & 0.601 & 0.378 & 0.622 \\ 
                \hline
            \end{tabular}
            }
    \end{center}
\end{table*}

\section{Conclusion}
We present a state-of-the-art framework for face swapping, MFIM. Our model adopts the GAN-inversion method using pretrained StyleGAN to generate a megapixel image and exploits 3DMM to supervise our model. Finally, we design a new operation, ID mixing, that creates a new identity using multiple source images and performs face swapping with that new identity.

However, the face swapping model can cause \textbf{negative impacts} on society. For example, a video made with a malicious purpose (e.g., fake news) can cause fatal damage to the victim. Nevertheless, it has positive impacts on the entertainment and theatrical industry. In addition, generating elaborate face-swapped images can contribute to advances in deepfake detection.


\clearpage

\renewcommand\thesection{Appendix.~\Alph{section}}
\setcounter{section}{0}

\section{Architecture}
\label{sup:section:architecture}
In this section, we describe the architectures of facial attribute encoder, generator and discriminator.

\subsection{Facial Attribute Encoder.}
\label{sup:subsection:facial_attribute_encoder}
Our facial attribute encoder, which is based on the psp~\cite{richardson2021encoding} encoder, uses the same encoder backbone (blue structures denoted as `Encoder Blocks' in Figure~\ref{subfig:face_swapping}) as the psp encoder. As shown in Figure~\ref{subfig:face_swapping}, the encoder backbone extracts the hierarchical latent maps from the given image. The M2C and M2M blocks of our facial attribute encoder extract the style codes and style maps from the hierarchical latent maps extracted from the backbone, respectively. The details of encoding process are as follows.

\subsubsection{Style codes.}
\label{sup:subsubsection:style_codes}
The architecture of the M2C block is the same as that of the Map2Style block of the pSp encoder. However, the pSp encoder produces eighteen style codes because it maps the image to $\mathcal{W^+}$ space~\cite{abdal2019image2stylegan}, whereas our facial attribute encoder maps the image to $\mathcal{S}$ space~\cite{wu2021stylespace}, so twenty-six style codes, $\left\{ c_i \right\}_{i=0}^{25}$. Then, the style codes go through the following additional steps:

\begin{gather}
    \label{eq:style_codes_transformation}
    s_i = \alpha_i c_i + \mu_i,
\end{gather}
where $\left\{ \alpha_i \right\}_{i=0}^{25}$ is a set of learnable parameters and $\left\{ \mu_i \right\}_{i=0}^{25}$ is a set of style codes that maps an average latent code of $\mathcal{W}$ space~\cite{karras2019style} to $\mathcal{S}$ space. $\alpha_i$, $c_i$, and $\mu_i$ have the same dimensions. 

We extract the style codes from the source image, $x_{src}$, and the target image, $x_{tgt}$, respectively, and combine them to construct the final style codes. Let us denote the style codes extracted from $x_{src}$ and $x_{tgt}$, $\left\{ s_i^{src} \right\}_{i=0}^{25}$ and $\left\{ s_i^{tgt} \right\}_{i=0}^{25}$, respectively. We construct the ID-irrelevant style codes, $\left\{ s_i^{tgt} \right\}_{i=0}^{b-1}$, by taking a subset of $\left\{ s_i^{tgt} \right\}_{i=0}^{25}$, and the ID style codes $\left\{ s_i^{src} \right\}_{i=b}^{25}$ from $\left\{ s_i^{src} \right\}_{i=0}^{25}$, where $b$ is a hyperparameter for the border index between the ID and ID-irrelevant style codes. We set $b=8$. Then, the final style codes, $\left\{ s_i \right\}_{i=0}^{25}$, are constructed by combining $\left\{ s_i^{tgt} \right\}_{i=0}^{b-1}$ and $\left\{ s_i^{src} \right\}_{i=b}^{25}$. Finally, $\left\{ s_i \right\}_{i=0}^{25}$ is used in weight demodulation operation~\cite{karras2020analyzing}. 

\subsubsection{Style maps.}
\label{sup:subsubsection:style_maps}
Our facial attribute encoder introduces an M2M block with the architecture depicted in Table~\ref{table:m2m_block} to extract the style maps from the target image. As shown in Table~\ref{table:m2m_block}, the M2M block takes the latent maps as input and produces two groups of style maps, which are denoted as \textbf{Output 0} and \textbf{Output 1} in Table~\ref{table:m2m_block} respectively, of the same spatial size as the input latent maps. 

Our encoder produces a total of \textit{four} groups of style maps: two groups with a spatial size of $16 \times 16$, $\left\{ m_i^{16 \times 16} \right\}_{i=0}^1$, and the remaining two groups have a spatial size of $32 \times 32$, $\left\{ m_i^{32 \times 32} \right\}_{i=0}^1$. All of these style maps have the channel dimensions of 512. Finally, these style maps are given to the pretrained StyleGAN generator as noise inputs.

\begin{table*}[t]
\centering
\caption{\textbf{Architecture of M2M block.} M2M block has shared convolutional layers at the top, but separated convolutional layers at the bottom. All convolutional layers have kernel size of $3\times3$, stride of 1, and padding size  of 1. $C_{in}$ and $C_{out}$ for the convolutional layer denotes the input and output channel dimensions, respectively. $a$ for the LeakyReLU layer denotes the negative slope. To encourage the style maps to be similar to the noise inputs which is used in StyleGAN pretraining, M2M block has an instance normalization~\cite{ulyanov2016instance} layer at the last which makes the style maps to be normally distributed.
}
    \label{table:m2m_block}
    \setlength{\tabcolsep}{0.7em} 
    \begin{tabular}{c|c}
    \multicolumn{2}{c}{(\textbf{Input}): latent maps ($c,h,w$)  } \\
    \toprule
    \multicolumn{2}{c}{Conv ($C_{in}=c$, $C_{out}=c$)} \\
    \multicolumn{2}{c}{LeakyReLU ($a = 0.01$)}\\
    \midrule
    \multicolumn{2}{c}{Conv ($C_{in}=c$, $C_{out}=c'$)} \\
    \multicolumn{2}{c}{LeakyReLU ($a = 0.01$)}\\
    \midrule
    Conv ($C_{in}=c'$, $C_{out}=c'$) & Conv ($C_{in}=c'$, $C_{out}=c'$) \\
    InstanceNorm & InstanceNorm \\
    \bottomrule
    (\textbf{Output 0}): style maps ($c',h,w$) & (\textbf{Output 1}): style maps ($c',h,w$) \\
    \end{tabular}
\end{table*}

\begin{table}
\centering
\caption{\textbf{Inputs that each layer of the pretrained StyleGAN generator takes for face swapping.} The ID-irrelevant style codes, $\left\{ s_i^{tgt} \right\}_{i=0}^{7}$, and style maps, $\left\{ m_i^{16 \times 16} \right\}_{i=0}^1$ and $\left\{ m_i^{32 \times 32} \right\}_{i=0}^1$ are extracted from $x_{tgt}$, while the ID style codes, $\left\{ s_i^{src} \right\}_{i=8}^{25}$ are extracted from $x_{src}$.}
    \label{table:inputs_face_swapping}
    \setlength{\extrarowheight}{0.3em}
    \setlength{\tabcolsep}{0.4em} 
    \begin{tabular}{cccccc}
    \toprule
    $\mathcal{S}$ layer index & Resolution & Layer name & Style code & Style code type & Style maps \\
    \midrule
    0 & $4 \times 4$ & Conv & $s_0^{tgt}$ & ID-irrelevant & - \\
    1 & $4 \times 4$ & ToRGB & $s_1^{tgt}$ & ID-irrelevant & - \\
    \midrule
    2 & $8 \times 8$ & ConvUp & $s_2^{tgt}$ & ID-irrelevant & - \\
    3 & $8 \times 8$ & Conv & $s_3^{tgt}$ & ID-irrelevant & - \\
    4 & $8 \times 8$ & ToRGB & $s_4^{tgt}$ & ID-irrelevant & - \\
    \midrule
    5 & $16 \times 16$ & ConvUp & $s_5^{tgt}$ & ID-irrelevant & $m_0^{16 \times 16}$ \\
    6 & $16 \times 16$ & Conv & $s_6^{tgt}$ & ID-irrelevant & $m_1^{16 \times 16}$ \\
    7 & $16 \times 16$ & ToRGB & $s_7^{tgt}$ & ID-irrelevant & - \\
    \midrule
    8 & $32 \times 32$ & ConvUP & $s_8^{src}$ & ID & $m_0^{32 \times 32}$ \\
    9 & $32 \times 32$ & Conv & $s_9^{src}$ & ID & $m_1^{32 \times 32}$ \\
    10 & $32 \times 32$ & ToRGB & $s_{10}^{src}$ & ID & - \\
    \midrule
    11 & $64 \times 64$ & ConvUP & $s_{11}^{src}$ & ID & - \\
    12 & $64 \times 64$ & Conv & $s_{12}^{src}$ & ID & - \\
    13 & $64 \times 64$ & ToRGB & $s_{13}^{src}$ & ID & - \\
    \midrule
    14 & $128 \times 128$ & ConvUP & $s_{14}^{src}$ & ID & - \\
    15 & $128 \times 128$ & Conv & $s_{15}^{src}$ & ID & - \\
    16 & $128 \times 128$ & ToRGB & $s_{16}^{src}$ & ID & - \\
    \midrule
    17 & $256 \times 256$ & ConvUP & $s_{17}^{src}$ & ID & - \\
    18 & $256 \times 256$ & Conv & $s_{18}^{src}$ & ID & - \\
    19 & $256 \times 256$ & ToRGB & $s_{19}^{src}$ & ID & - \\
    \midrule
    20 & $512 \times 512$ & ConvUP & $s_{20}^{src}$ & ID & - \\
    21 & $512 \times 512$ & Conv & $s_{21}^{src}$ & ID & - \\
    22 & $512 \times 512$ & ToRGB & $s_{22}^{src}$ & ID & - \\
    \midrule
    23 & $1024 \times 1024$ & ConvUP & $s_{23}^{src}$ & ID & - \\
    24 & $1024 \times 1024$ & Conv & $s_{24}^{src}$ & ID & - \\
    25 & $1024 \times 1024$ & ToRGB & $s_{25}^{src}$ & ID & - \\
    \bottomrule
    \end{tabular}
\end{table}

\begin{table}
\centering
\caption{\textbf{Inputs that each layer of the pretrained StyleGAN generator takes for ID mixing.} The ID-irrelevant style codes, $\left\{ s_i^{tgt} \right\}_{i=0}^{7}$, and style maps, $\left\{ m_i^{16 \times 16} \right\}_{i=0}^1$ and $\left\{ m_i^{32 \times 32} \right\}_{i=0}^1$ are extracted from $x_{tgt}$. However, the global ID style codes, $\left\{ s_i^{src} \right\}_{i=8}^{9}$, are extracted from $x_{src}^{gb}$, and the local ID style codes, $\left\{ s_i^{src} \right\}_{i=10}^{25}$, are extracted from $x_{src}^{lc}$.}

    \label{table:inputs_id_mixing}
    \setlength{\extrarowheight}{0.3em}
    \setlength{\tabcolsep}{0.4em} 
    \begin{tabular}{cccccc}
    \toprule
    $\mathcal{S}$ layer index & Resolution & Layer name & Style code & Style code type & Style maps \\
    \midrule
    0 & $4 \times 4$ & Conv & $s_0^{tgt}$ & ID-irrelevant & - \\
    1 & $4 \times 4$ & ToRGB & $s_1^{tgt}$ & ID-irrelevant & - \\
    \midrule
    2 & $8 \times 8$ & ConvUp & $s_2^{tgt}$ & ID-irrelevant & - \\
    3 & $8 \times 8$ & Conv & $s_3^{tgt}$ & ID-irrelevant & - \\
    4 & $8 \times 8$ & ToRGB & $s_4^{tgt}$ & ID-irrelevant & - \\
    \midrule
    5 & $16 \times 16$ & ConvUp & $s_5^{tgt}$ & ID-irrelevant & $m_0^{16 \times 16}$ \\
    6 & $16 \times 16$ & Conv & $s_6^{tgt}$ & ID-irrelevant & $m_1^{16 \times 16}$ \\
    7 & $16 \times 16$ & ToRGB & $s_7^{tgt}$ & ID-irrelevant & - \\
    \midrule
    8 & $32 \times 32$ & ConvUP & $s_8^{src}$ & Global ID & $m_0^{32 \times 32}$ \\
    9 & $32 \times 32$ & Conv & $s_9^{src}$ & Global ID & $m_1^{32 \times 32}$ \\
    10 & $32 \times 32$ & ToRGB & $s_{10}^{src}$ & Local ID & - \\
    \midrule
    11 & $64 \times 64$ & ConvUP & $s_{11}^{src}$ & Local ID & - \\
    12 & $64 \times 64$ & Conv & $s_{12}^{src}$ & Local ID & - \\
    13 & $64 \times 64$ & ToRGB & $s_{13}^{src}$ & Local ID & - \\
    \midrule
    14 & $128 \times 128$ & ConvUP & $s_{14}^{src}$ & Local ID & - \\
    15 & $128 \times 128$ & Conv & $s_{15}^{src}$ & Local ID & - \\
    16 & $128 \times 128$ & ToRGB & $s_{16}^{src}$ & Local ID & - \\
    \midrule
    17 & $256 \times 256$ & ConvUP & $s_{17}^{src}$ & Local ID & - \\
    18 & $256 \times 256$ & Conv & $s_{18}^{src}$ & Local ID & - \\
    19 & $256 \times 256$ & ToRGB & $s_{19}^{src}$ & Local ID & - \\
    \midrule
    20 & $512 \times 512$ & ConvUP & $s_{20}^{src}$ & Local ID & - \\
    21 & $512 \times 512$ & Conv & $s_{21}^{src}$ & Local ID & - \\
    22 & $512 \times 512$ & ToRGB & $s_{22}^{src}$ & Local ID & - \\
    \midrule
    23 & $1024 \times 1024$ & ConvUP & $s_{23}^{src}$ & Local ID & - \\
    24 & $1024 \times 1024$ & Conv & $s_{24}^{src}$ & Local ID & - \\
    25 & $1024 \times 1024$ & ToRGB & $s_{25}^{src}$ & Local ID & - \\
    \bottomrule
    \end{tabular}
\end{table}

\subsection{Generator}
\label{sup:subsection:generator}
We use the pretrained generator of StyleGAN~\cite{karras2020analyzing}, so we use the same architecture with StyleGAN without modification except for the mapping network that maps a random vector $z \in \mathcal{Z}$ to an intermediate latent space $\mathcal{W}$. We replace the mapping network with the facial attribute encoder which produces the ID-irrelevant style codes, ID style codes and style maps. These are forwarded appropriately to each layer of the pretrained StyleGAN generator, as shown in Tables~\ref{table:inputs_face_swapping} and~\ref{table:inputs_id_mixing}. Table~\ref{table:inputs_face_swapping} describes the process of face swapping, which uses a single source image, $x_{src}$, but Table~\ref{table:inputs_id_mixing} describes the process of id mixing, which uses the global and local source images, $x_{src}^{gb}$ and $x_{src}^{lc}$.

\subsection{Discriminator}
\label{sup:subsection:discriminator}
We use the pretrained discriminator of StyleGAN~\cite{karras2020analyzing}, so we use the same architecture with StyleGAN without modification.

\section{Hyperparameters}
\label{sec:hyperparameters}
Table~\ref{table:hyperparameters_lambda} shows weights for each loss to train our model. Following StyleGAN~\cite{karras2020analyzing}, we use R1 regularization~\cite{mescheder2018training} every sixteen training steps. Table~\ref{table:hyperparameters_optimization} shows additional hyperparameters for optimization. 
For the optimizer, we use the Ranger optimizer, which is a combination of RAdam~\cite{liu2019radam} and Lookahead~\cite{zhang2019lookahead}, following pSp~\cite{richardson2021encoding}. We use a learning rate of $1e-4$ and decrease it by $2e-5$ every 40,000 steps after 500,000 steps.
We use a batch size of four, which means that we use four pairs of source and target images for training. However, for one of the four pairs, we make the source image and the target image the same, so that the generator performs self-reconstruction on that pair.

\begin{table}
\centering
\caption{\textbf{Weights for each loss.} Each loss is described in the main manuscript.}
    \label{table:hyperparameters_lambda}
    \setlength{\tabcolsep}{0.4em} 
    \begin{tabular}{cccccccc}
    \toprule
    $\lambda_{id}$ & $\lambda_{recon}$ & $\lambda_{adv}$ & $\lambda_{R_1}$ & $\lambda_{shape}$ & $\lambda_{pose}$ & $\lambda_{exp}$ & $R_1$ step \\
    \midrule
    2.0 & 1.0 & 0.1 & 10.0 & 5.0 & 1.0 & 1.0 & 16 \\
    \bottomrule
    \end{tabular}
\end{table}

\begin{table}
\centering
\caption{\textbf{Hyperparameters for optimization.} The details are described in Section~\ref{sec:hyperparameters}.}
    \label{table:hyperparameters_optimization}
    \setlength{\tabcolsep}{0.2em} 
    \begin{tabular}{cccccc}
    \toprule
    Training steps & Optimizer & Learning rate & Learning rate decay & Batch size & Self-recon size \\
    \midrule
    700,000 & Ranger & 0.0001 & Step & 4 & 1 \\
    \bottomrule
    \end{tabular}
\end{table}

\section{Preprocess and Postprocess}
\subsection{Data preprocess}
We use FFHQ~\cite{karras2019style}, which consists of 70,000 human faces at $1024 \times 1024$ resolution, for the training dataset. It is noteworthy that the most of the previous face-swapping models~\cite{li2019faceshifter,wang2021hififace,zhu2021one,gao2021information} extend the training dataset by combining multiple datasets, but we only use FFHQ. Therefore, our model can be trained more efficiently because our model does not require any additional preprocess steps such as image alignment to combine the multiple datasets.

For training, we basically follow the image preprocess protocol of pSp~\cite{richardson2021encoding}. However, for $\mathcal{L}_{adv}$ and $R_1$, we use the images with the size of $1024 \times 1024$. Furthermore, for 3DMM supervision, we preprocess the images by following the image preprocess protocol of DECA~\cite{DECA:Siggraph2021} before forwarding the images to DECA encoder. 

\subsection{Postprocess: ROI Only Synthesis}
\label{subsec:roi_only_synthesis}
Our model can faithfully reconstruct the background or hair style of $x_{tgt}$, but we can further improve our model to reconstruct the high-frequency details of the background or hair style via ROI only synthesis. 

\begin{figure*}[t]
    \begin{center}
    \includegraphics[width=1\linewidth]{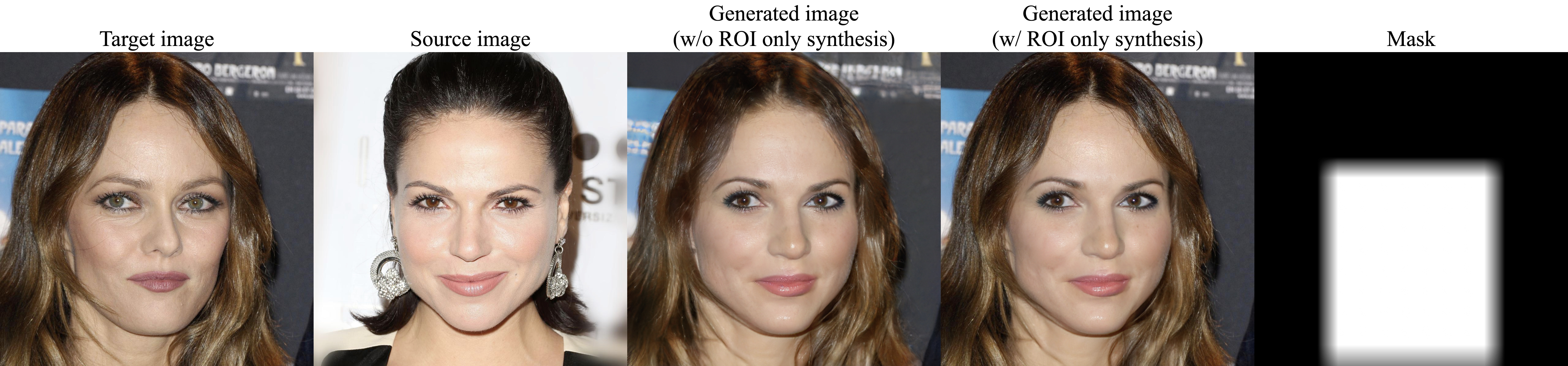}
    \end{center}
    \caption{\textbf{ROI only synthesis.} The details are described in Section~\ref{subsec:roi_only_synthesis}.}
    \label{fig:roi_only_synthesis}
\end{figure*}

Note that it does not require a segmentation label at all. This process is depicted in Figure~\ref{fig:roi_only_synthesis}. Assuming that the image is aligned, we use a mask, which has a size of $1024 \times 1024$, with a fixed box at the expected location of the face. Specifically, we set the size of the box to a width of 512 and a height of 608 and top-left coordinates, $(top,~left)$, to $(384,~256)$. The inside of the box has a value of one, and the outside has a value of zero. Then, we blur the boundary by downsampling the mask to the size of $16 \times 16$ and upsampling it to the size of $1024 \times 1024$ again. With this mask, the final output image is generated as 

\begin{gather}
    \label{eq:roi_only_synthesis}
    m \odot x_{swap} + (1-m) \odot x_{tgt},
\end{gather}
where $m$ is a mask and $\odot$ is the element-wise product. Note that it is not used at the training phase, only at the inference phase. Also, we use it only in the qualitative results, not in the quantitative results at all.

\section{Analysis on 3DMM Supervision}
\label{subsec:sup_3dmm_supervision}
We compare our 3DMM supervision method and that of HifiFace~\cite{wang2021hififace}. We first describe each method and then compare them with experimental results.

\subsection{Method }
For the 3DMM supervision, our model utilizes the 3DMM parameter reconstruction loss which is formulated as
\begin{gather}
    \label{eq:parameter_recon_loss}
    \mathcal{L}_{param} = \lambda_{shape} \mathcal{L}_{shape} + \lambda_{pose} \mathcal{L}_{pose} + \lambda_{exp} \mathcal{L}_{exp},
\end{gather}
where $\mathcal{L}_{shape}$, $\mathcal{L}_{pose}$, and $\mathcal{L}_{exp}$ are described in the main manuscript, and $\lambda_{shape}$, $\lambda_{pose}$, and $\lambda_{exp}$ are weights for each loss. 

On the other hand, HifiFace utilizes the landmark reconstruction loss. Note that 3DMM can reconstruct a 3D face using 3DMM parameters and extract landmark keypoints corresponding to the 3D face. Using this capability, HifiFace encourages the landmark keypoints of the generated image, $\left\{ q^{gen}_k \right\}_{k=1}^K$ to be equal to its ground-truth landmark keypoints, $\left\{ q^{gt}_k \right\}_{k=1}^K$. Here, when using DECA~\cite{DECA:Siggraph2021}, $K=68$, and the ground-truth landmark keypoints are extracted from the reconstructed 3D face using the shape parameter of the source image and the pose, expression, and cam parameters of the target image. We apply this method to our model to formulate the landmark reconstruction loss as
\begin{gather}
    \label{eq:landmark_recon_loss}
    \mathcal{L}_{lm} = \frac{1}{K} \sum_{k=1}^{K} || \left\{ q^{gen}_k \right\} - \left\{ q^{gt}_k \right\} ||_1.
\end{gather}

\begin{table*}[t]
    \begin{center}
    \caption{\textbf{Quantitative comparison between $\mathcal{L}_{lm}$ and $\mathcal{L}_{param}$.} The metrics are the same with those in the main manuscript. Also, the configuration (B) is the same with that in the main manuscript. $\mathcal{L}_{lm}$ and $\mathcal{L}_{param}$ are the 3DMM supervision methods of HifiFace~\cite{wang2021hififace} and ours, respectively.}
    \label{table:ablation_study_3dmm_supervision}
        \setlength{\tabcolsep}{0.7em} 
        \scalebox{1}{
            \begin{tabular}{l|ccccc}
                \hline\noalign{\smallskip}
                $~~~~$Configuration & Identity $\downarrow$ & Shape $\downarrow$ & Expression $\downarrow$ & Pose $\downarrow$ & Pose-HN $\downarrow$ \\
                \hline\noalign{\smallskip}
                \textrm{B}. & 96.066 & 0.887 & 0.424 & 0.053 & 3.839 \\
                \textrm{B} $+~\mathcal{L}_{lm}$. & 96.016 & 0.892 & 0.418 & 0.046 & 3.683 \\
                \textrm{B} $+~\mathcal{L}_{param}$. & 96.153 & 0.842 & 0.426 & 0.060 & 4.173 \\
                \hline
            \end{tabular}
            }
    \end{center}
\end{table*}

\begin{table*}
    \begin{center}
    \caption{\textbf{Full ablation study of MFIM.} This table is the same with the table of the ablation study in the main manuscript, but the configuration (F) is newly added.}
    \label{table:ablation_study_full}
        \setlength{\tabcolsep}{0.7em} 
        \scalebox{1}{
            \begin{tabular}{l|ccccc}
                \hline\noalign{\smallskip}
                $~~~~$Configuration & Identity $\downarrow$ & Shape $\downarrow$ & Expression $\downarrow$ & Pose $\downarrow$ & Pose-HN $\downarrow$ \\
                \hline\noalign{\smallskip}
                \textrm{A}. Baseline MFIM & 70.160 & 0.383 & 1.116 & 0.145 & 7.899 \\
                \textrm{B}. $+$ style maps & 91.430 & 0.823 & 0.398 & 0.051 & 3.795 \\
                \textrm{C}. $+~\mathcal{L}_{shape}$ & 86.476 & 0.635 & 0.864 & 0.085 & 5.091 \\
                \textrm{D}. $+~\mathcal{L}_{pose}$ & 86.777 & 0.634 & 0.860 & 0.078 & 4.797 \\
                \textrm{E}. $+~\mathcal{L}_{exp}$ & 91.469 & 0.782 & 0.400 & 0.057 & 4.095 \\
                \textrm{F}. $+~\mathcal{L}_{lm}$ & 92.018 & 0.778 & 0.387 & 0.041 & 3.876 \\
                \hline
            \end{tabular}
            }
    \end{center}
\end{table*}

\subsection{Comparison between $\mathcal{L}_{lm}$ and $\mathcal{L}_{param}$ }
In Table~\ref{table:ablation_study_3dmm_supervision}, we compare $\mathcal{L}_{lm}$ and $\mathcal{L}_{param}$ on CelebA-HQ~\cite{karras2017progressive}. Here, unlike the quantitative experiment on CelebA-HQ in the main manuscript, we use 30,000 face-swapped images instead of 300,000. Specifically, we randomly assign an image to each image in CelebA-HQ and make 30,000 pairs of the source image and target image. 

The configuration (B) in Table~\ref{table:ablation_study_3dmm_supervision} is the same with that in the main manuscript. Then, we construct the configurations (B+$\mathcal{L}_{lm}$) and (B+$\mathcal{L}_{param}$) by adding $\mathcal{L}_{lm}$ and $\mathcal{L}_{parm}$ to the configuration (B), respectively. The configuration (B+$\mathcal{L}_{param}$) is the same with the configuration (E), our proposed model, in the main manuscript.

\begin{figure*}
    \begin{center}
    \includegraphics[width=1\linewidth]{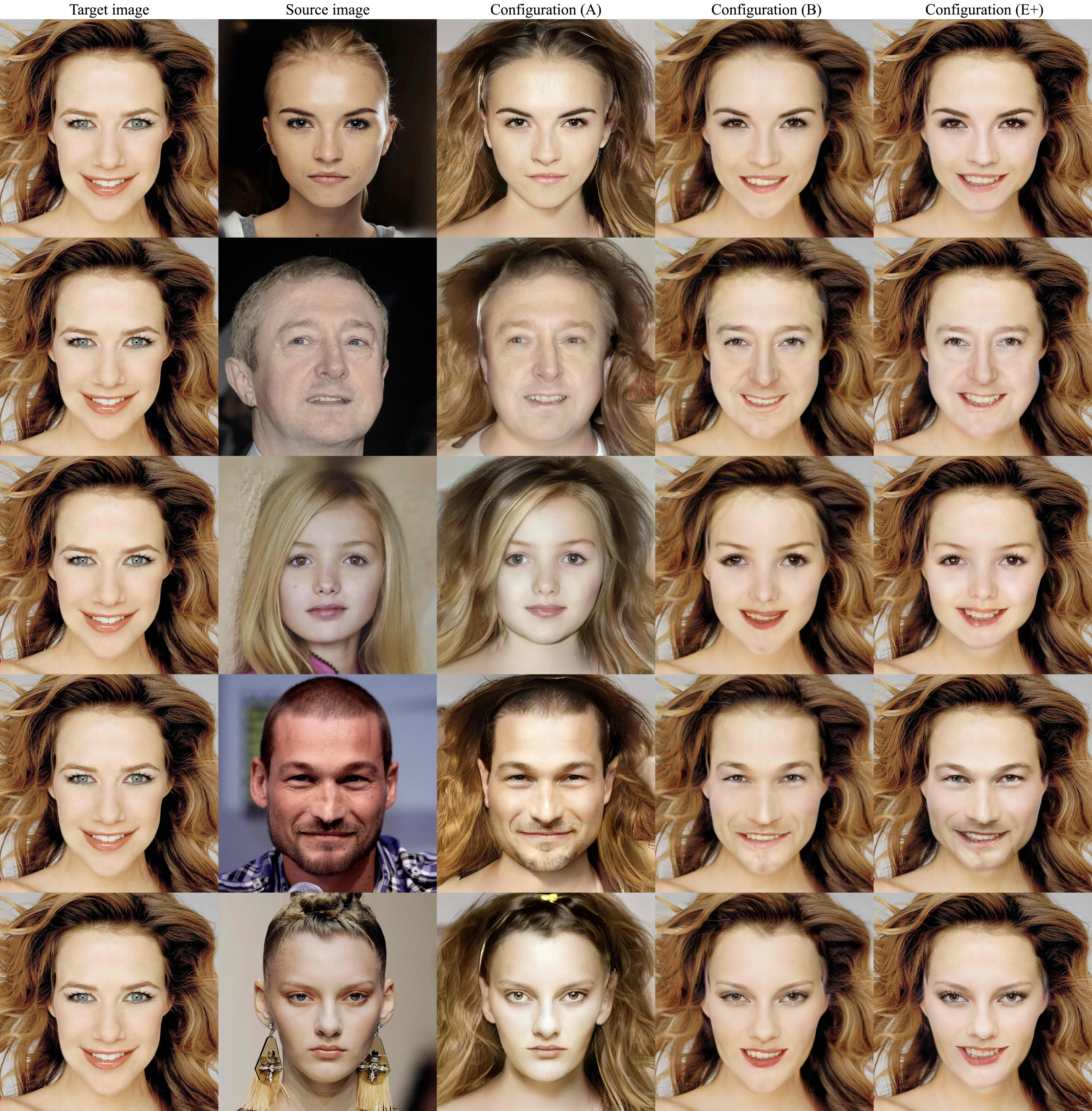}
    \end{center}
    \caption{\textbf{Ablation study of MFIM.} The configuration (A) transfers the ID attributes (e.g., eyes and face shape) of the source image while maintaining the overall structure and pose of the target image, but cannot reconstruct the details of the target image. The configuration (B) reconstructs the details of the target image better than the configuration (A), but the face shape of the source image is not sufficiently transferred. Finally, the configuration (E+) sufficiently transfers the face shape of the source image while preserving the ID-irrelevant attributes (e.g., pose and expression) of the target image. Furthermore, ROI only synthesis (Section~\ref{subsec:roi_only_synthesis}) allows our model to preserve the high-frequency details on hair or background of the target image.}
    \label{fig:sup_ablation_study}
\end{figure*}

As shown in Table~\ref{table:ablation_study_3dmm_supervision}, adding $\mathcal{L}_{lm}$ to the configuration (B) does not improve the shape score while $\mathcal{L}_{param}$ improves the shape score. However, we can see that $\mathcal{L}_{lm}$ improves the pose score by comparing the configurations (B) and (B+$\mathcal{L}_{lm}$). We think that this may be because the pose, which is the more global attribute than the shape and expression, has a greater effect on the landmark regression than the shape or expression. For this reason, the most effective way to decrease $\mathcal{L}_{lm}$ can be to match the pose of the generated image to that of the target image. As a result, the model focuses on matching poses, and may not be sufficiently motivated to improve the shape score. 

In contrast, we use a separate loss for each attribute. In particular, to decrease $\mathcal{L}_{shape}$, the face shape of the generated image should be the same as that of the source image. Due to this difference, $\mathcal{L}_{param}$ can improve the shape score, while $\mathcal{L}_{lm}$ cannot. Although the pose score is somewhat degraded after applying $\mathcal{L}_{param}$, transforming the face shape rather than preserving the pose is one of our important goals. Furthermore, the configuration (B+$\mathcal{L}_{param}$) still shows the visually plausible results in terms of the pose. Therefore, we propose $\mathcal{L}_{param}$ as our 3DMM supervision method.

\begin{figure*}[t]
    \begin{center}
    \includegraphics[width=1\linewidth]{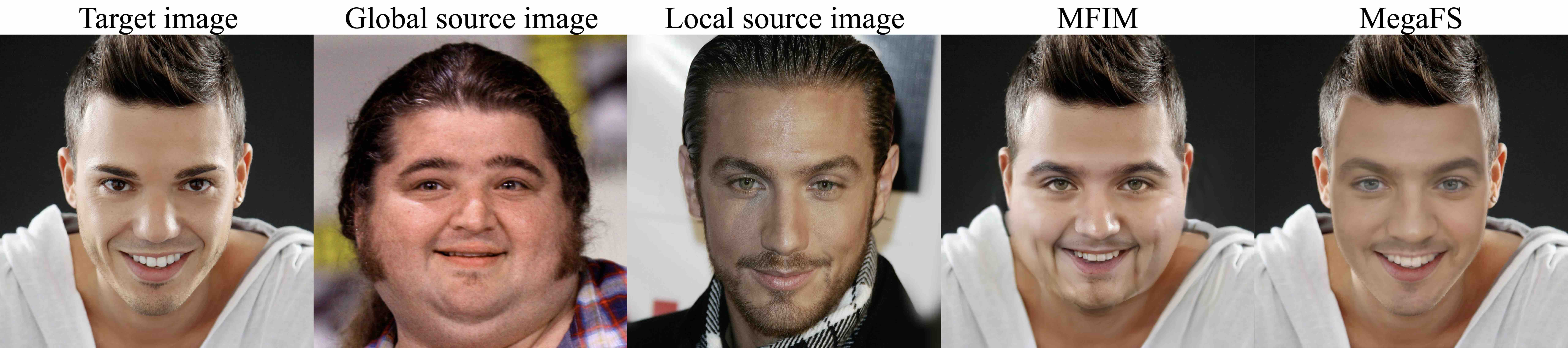}
    \end{center}
    \caption{\textbf{Qualitative comparison on ID mixing.} MegaFS has a trouble in ID mixing because it cannot transfer the face shape of the global source image. In contrast, our model can create a new identity by blending the global (e.g., face shape) and local (e.g., eyes) ID attributes captured from the global and local source images, respectively.}
    \label{fig:sup_id_mixing_comparison}
\end{figure*}

\subsection{Combination of $\mathcal{L}_{lm}$ and $\mathcal{L}_{param}$ }
Based on the results in Table~\ref{table:ablation_study_3dmm_supervision}, we further improve our model by combining $\mathcal{L}_{param}$ and $\mathcal{L}_{lm}$ as shown in Table~\ref{table:ablation_study_full}. For the results in Table~\ref{table:ablation_study_full}, we use 300,000 face-swapped images, which is the same setting with that of the quantitative experiment on CelebA-HQ in the main manuscript.

In Table~\ref{table:ablation_study_full}, we construct the configuration (F) by adding $\mathcal{L}_{lm}$ with the weight for this loss of 1,000 (i.e., $\lambda_{lm}=1000$) to the configuration (E). Here, we use only some of the landmark keypoints instead of the full landkark keypoints to encourage our model to further focus on matching the pose. Specifically, we use $\left\{ q^{gen}_k \right\}_{k \in \left\{9, 31, 37, 46, 49, 55 \right\}}$ and $\left\{ q^{gt}_k \right\}_{k \in \left\{9, 31, 37, 46, 49, 55 \right\}}$ . As shown in Table~\ref{table:ablation_study_full}, the configuration (F) achieves the better pose and pose-HN scores than the configuration (E) without deterioration on the shape and expression scores. As a result, the configuration (F) achieves the better shape, expression, and pose scores than the configuration (B) at the same time. However, $\mathcal{L}_{lm}$ is not our contribution and the configuration (E) also shows the visually plausible results in terms of pose, so we propose the configuration (E) as our final model.

Figure~\ref{fig:sup_ablation_study} shows the qualitative results for several configurations. We construct the configuration (E+) by adding ROI only synthesis~(Section~\ref{subsec:roi_only_synthesis}) to the configuration (E). As shown in Figure~\ref{fig:sup_ablation_study}, the configuration (E+) transfers the ID attributes (e.g., eyes and face shape) of the source image actively while preserving the ID-irrelevant attributes (e.g., pose and expression) of the target image. In Figure~\ref{fig:sup_ablation_study}, the differences between the configurations (A) and (B) show the effectiveness of the style maps, and the differences between the configurations (B) and (E+) show the effectiveness of the 3DMM supervision.

\section{Comparison with MegaFS on ID Mixing}
One of the state-of-the-art models, MegaFS~\cite{zhu2021one}, has a potential to perform ID mixing because it also exploits the StyleGAN~\cite{karras2020analyzing} architecture. However, MegaFS is not good at transforming the face shape as demonstrated in the manuscript. As a result, in Fig.~\ref{fig:sup_id_mixing_comparison}, MegaFS fails to performing ID mixing because it cannot transfer the round face shape of the global source image to the target image. It only transfers the eyes of the local source image to the target image. For this reason, the generated image by MegaFS does not seem an ID-mixed image. In contrast, our model, MFIM, can transfer the round face shape of the global source image and the eyes of the local source image at the same time. As a result, the generated image by MFIM seems an ID-mixed image.

\section{Additional Samples}
\begin{figure*}[t]
    \begin{center}
    \includegraphics[width=1\linewidth]{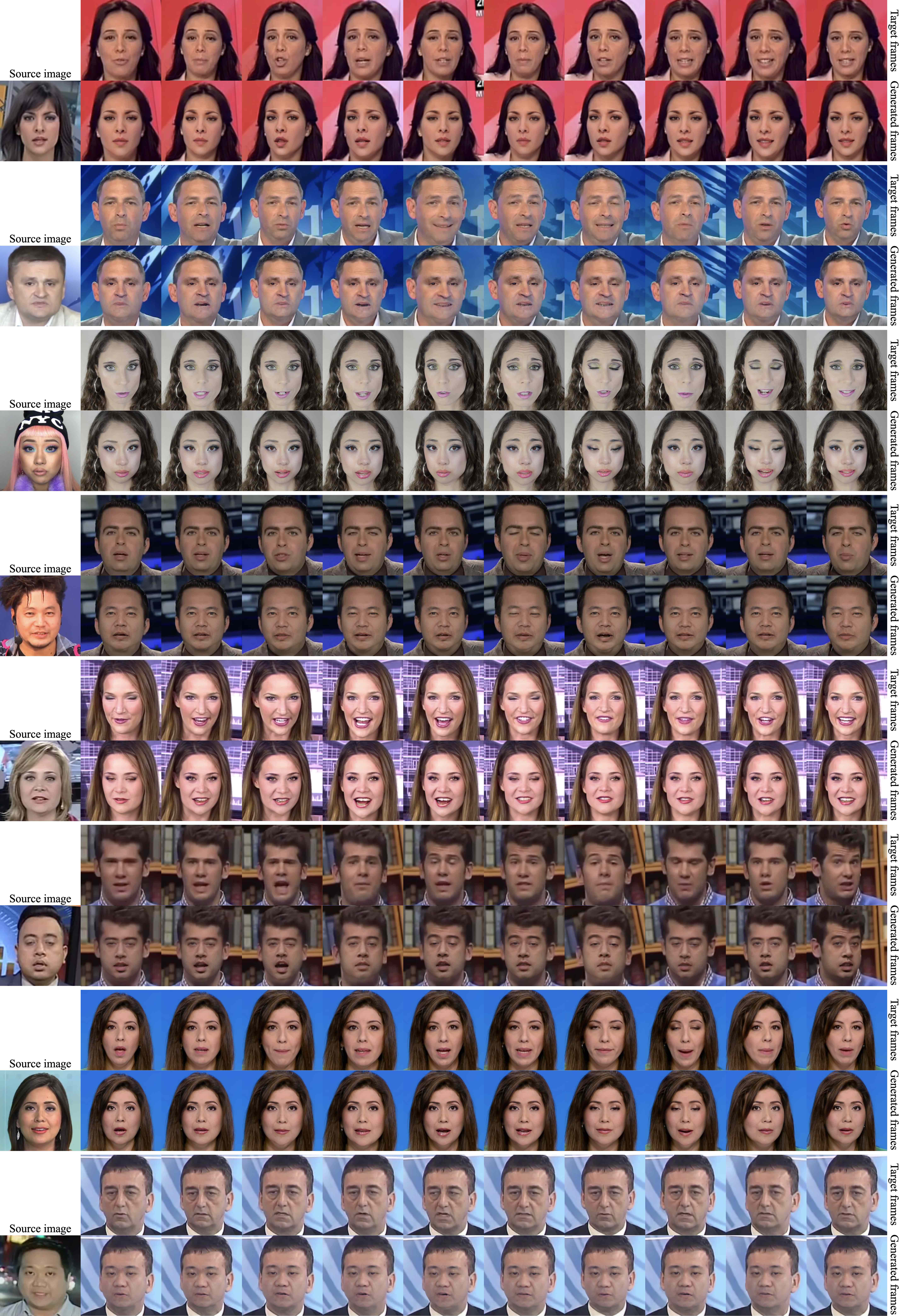}
    \end{center}
    \vskip -0.25in
    \caption{\textbf{Qualitative results of face swapping on FaceForensics++.} The leftmost image is the source image, and the uppermost images are the target frames captured uniformly from the video. The rest of the images are the generated frames.}
    \label{fig:sup_qualitative_result_ff}
\end{figure*}

\begin{figure*}[t]
    \begin{center}
    \includegraphics[width=1\linewidth]{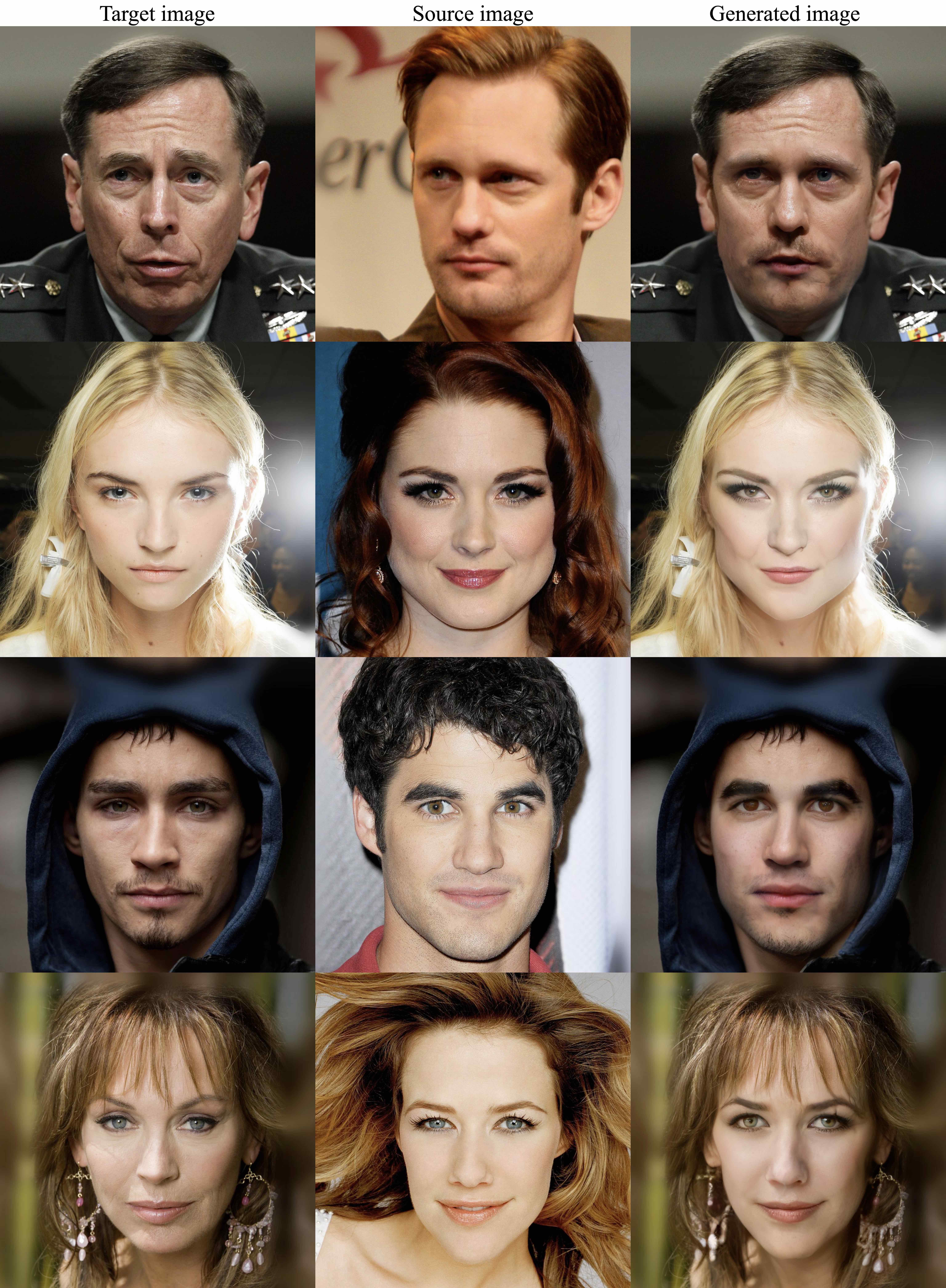}
    \end{center}
    \caption{\textbf{Qualitative results of face swapping on CelebA-HQ.} Our model faithfully captures ID (e.g., eyes and face shape) and ID-irrelevant (e.g., pose and expression) attributes from the source and target images, respectively, and synthesizes a high-quality megapixel image by blending these attributes.}
    \label{fig:sup_qualitative_result_celeba_hq_1}
\end{figure*}

\begin{figure*}[t]
    \begin{center}
    \includegraphics[width=1\linewidth]{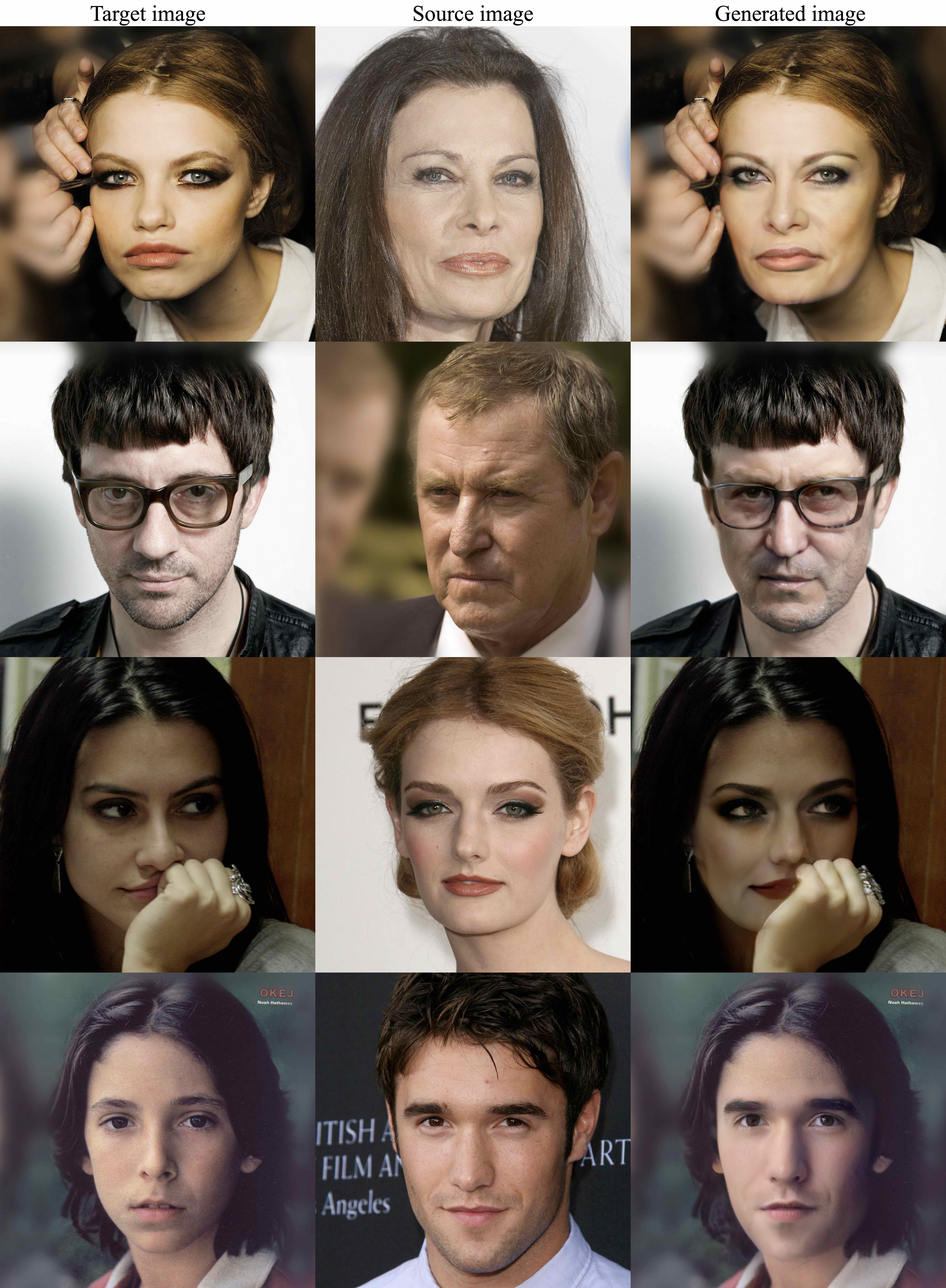}
    \end{center}
    \caption{\textbf{Qualitative results of face swapping on CelebA-HQ.} Our model faithfully captures ID (e.g., eyes and face shape) and ID-irrelevant (e.g., pose and expression) attributes from the source and target images, respectively, and synthesizes a high-quality megapixel image by blending these attributes.}
    \label{fig:sup_qualitative_result_celeba_hq_2}
\end{figure*}

\begin{figure*}[t]
    \begin{center}
    \includegraphics[width=1\linewidth]{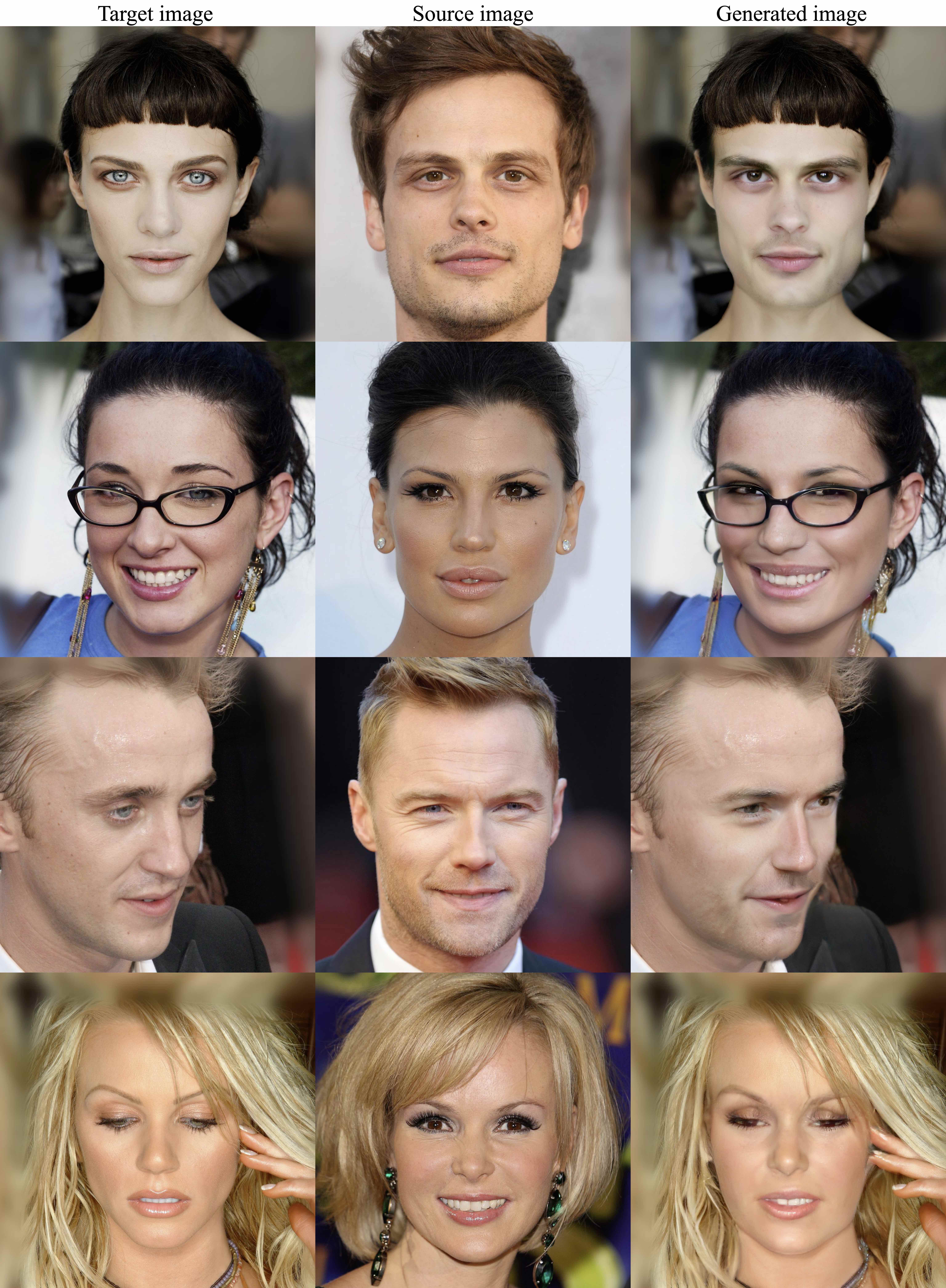}
    \end{center}
    \caption{\textbf{Qualitative results of face swapping on CelebA-HQ.} Our model faithfully captures ID (e.g., eyes and face shape) and ID-irrelevant (e.g., pose and expression) attributes from the source and target images, respectively, and synthesizes a high-quality megapixel image by blending these attributes.}
    \label{fig:sup_qualitative_result_celeba_hq_hard}
\end{figure*}

\begin{figure*}[t]
    \begin{center}
    \includegraphics[width=1\linewidth]{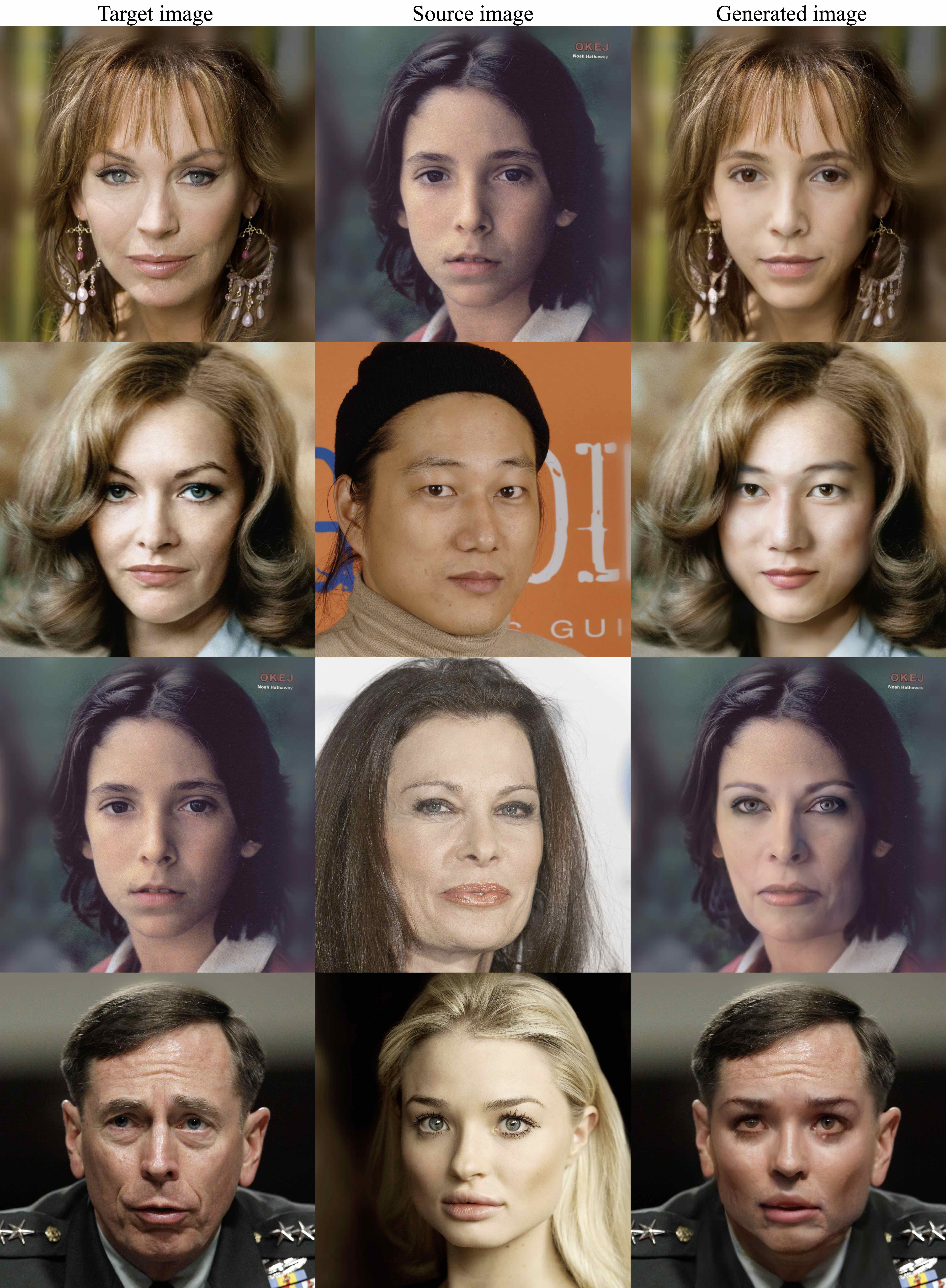}
    \end{center}
    \caption{\textbf{Qualitative results of large-gap face swapping on CelebA-HQ.} Our model faithfully performs face swapping even with a large gap between the source and target images (e.g., gender and age).}
    \label{fig:sup_qualitative_result_celeba_hq_large_gap}
\end{figure*}

\begin{figure*}[t]
    \begin{center}
    \includegraphics[width=1\linewidth]{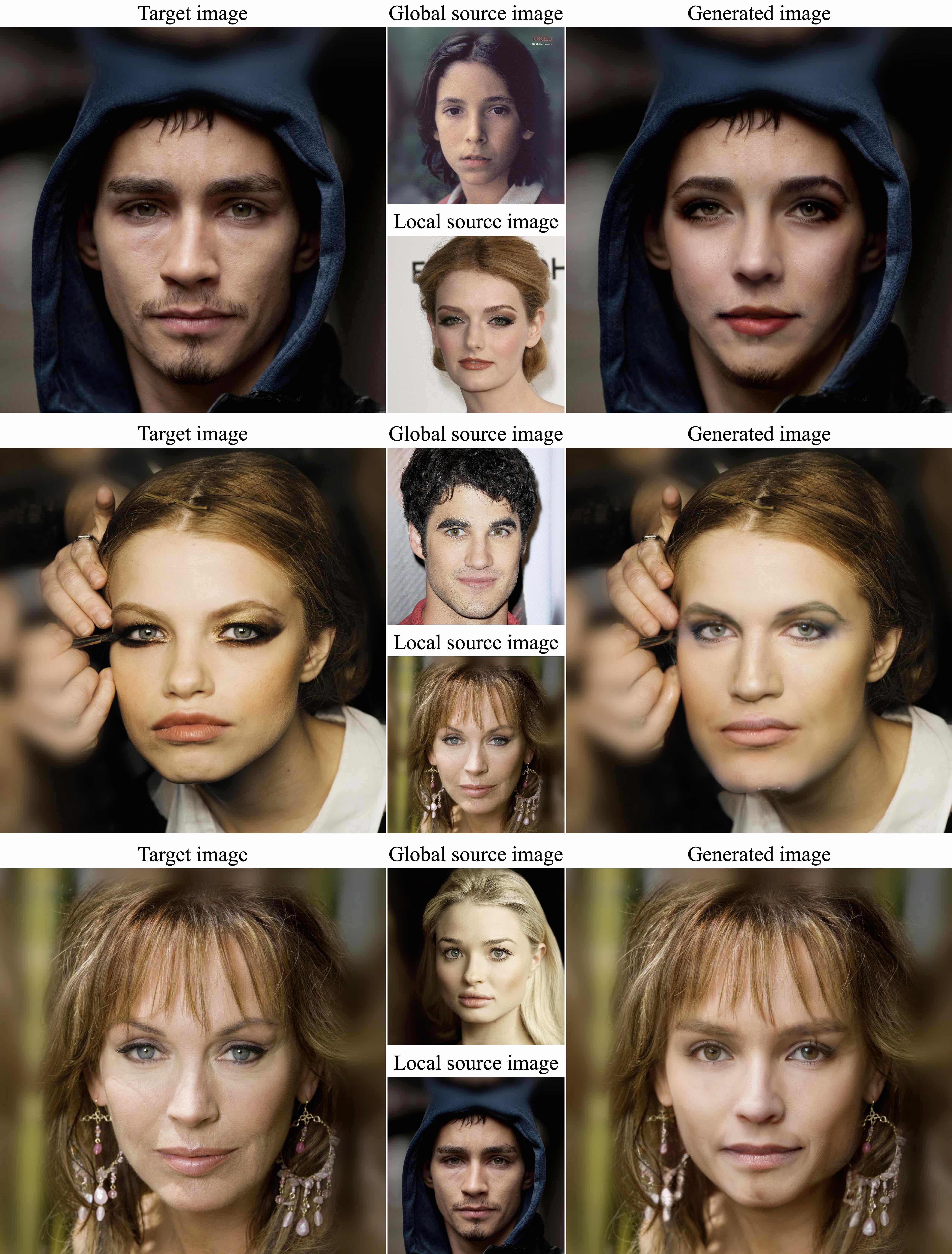}
    \end{center}
    \caption{\textbf{Qualitative results of ID mixing on CelebA-HQ.} Our model can create a new identity by blending the global (e.g., face shape) and local (e.g., eyes) ID attributes captured from the global and local source images, respectively.}
    \label{fig:sup_qualitative_result_id_mixing}
\end{figure*}

Figure~\ref{fig:sup_qualitative_result_ff} shows the qualitative results of face swapping on FaceForensics++~\cite{rossler2019faceforensics++}. Figures~\ref{fig:sup_qualitative_result_celeba_hq_1}, \ref{fig:sup_qualitative_result_celeba_hq_2},~\ref{fig:sup_qualitative_result_celeba_hq_hard}, and \ref{fig:sup_qualitative_result_celeba_hq_large_gap} show the qualitative results  of face swapping on CelebA-HQ~\cite{karras2017progressive}. Figure~\ref{fig:sup_qualitative_result_id_mixing} shows the qualitative results of ID mixing on CelebA-HQ.

\clearpage
\bibliographystyle{splncs04}
\bibliography{egbib}

\begin{thebibliography}{10}
\providecommand{\url}[1]{\texttt{#1}}
\providecommand{\urlprefix}{URL }
\providecommand{\doi}[1]{https://doi.org/#1}

\bibitem{deepfakes}
Deepfakes.
  \url{https://github.com/ondyari/FaceForensics/tree/master/dataset/DeepFakes}

\bibitem{abdal2019image2stylegan}
Abdal, R., Qin, Y., Wonka, P.: Image2stylegan: How to embed images into the
  stylegan latent space? In: Proceedings of the IEEE/CVF International
  Conference on Computer Vision. pp. 4432--4441 (2019)

\bibitem{alaluf2021restyle}
Alaluf, Y., Patashnik, O., Cohen-Or, D.: Restyle: A residual-based stylegan
  encoder via iterative refinement. In: Proceedings of the IEEE/CVF
  International Conference on Computer Vision. pp. 6711--6720 (2021)

\bibitem{alexander2009digital}
Alexander, O., Rogers, M., Lambeth, W., Chiang, M., Debevec, P.: The digital
  emily project: photoreal facial modeling and animation. In: Acm siggraph 2009
  courses, pp. 1--15 (2009)

\bibitem{bahng2020exploring}
Bahng, H., Chung, S., Yoo, S., Choo, J.: Exploring unlabeled faces for novel
  attribute discovery. In: Proceedings of the IEEE/CVF Conference on Computer
  Vision and Pattern Recognition. pp. 5821--5830 (2020)

\bibitem{blanz1999morphable}
Blanz, V., Vetter, T.: A morphable model for the synthesis of 3d faces. In:
  Proceedings of the 26th annual conference on Computer graphics and
  interactive techniques. pp. 187--194 (1999)

\bibitem{brock2018large}
Brock, A., Donahue, J., Simonyan, K.: Large scale gan training for high
  fidelity natural image synthesis. arXiv preprint arXiv:1809.11096  (2018)

\bibitem{cao2013facewarehouse}
Cao, C., Weng, Y., Zhou, S., Tong, Y., Zhou, K.: Facewarehouse: A 3d facial
  expression database for visual computing. IEEE Transactions on Visualization
  and Computer Graphics  \textbf{20}(3),  413--425 (2013)

\bibitem{cao2018vggface2}
Cao, Q., Shen, L., Xie, W., Parkhi, O.M., Zisserman, A.: Vggface2: A dataset
  for recognising faces across pose and age. In: 2018 13th IEEE international
  conference on automatic face \& gesture recognition (FG 2018). pp. 67--74.
  IEEE (2018)

\bibitem{chen2020simswap}
Chen, R., Chen, X., Ni, B., Ge, Y.: Simswap: An efficient framework for high
  fidelity face swapping. In: Proceedings of the 28th ACM International
  Conference on Multimedia. pp. 2003--2011 (2020)

\bibitem{cho2019image}
Cho, W., Choi, S., Park, D.K., Shin, I., Choo, J.: Image-to-image translation
  via group-wise deep whitening-and-coloring transformation. In: Proceedings of
  the IEEE/CVF Conference on Computer Vision and Pattern Recognition. pp.
  10639--10647 (2019)

\bibitem{choi2018stargan}
Choi, Y., Choi, M., Kim, M., Ha, J.W., Kim, S., Choo, J.: Stargan: Unified
  generative adversarial networks for multi-domain image-to-image translation.
  In: Proceedings of the IEEE conference on computer vision and pattern
  recognition. pp. 8789--8797 (2018)

\bibitem{choi2020stargan}
Choi, Y., Uh, Y., Yoo, J., Ha, J.W.: Stargan v2: Diverse image synthesis for
  multiple domains. In: Proceedings of the IEEE/CVF conference on computer
  vision and pattern recognition. pp. 8188--8197 (2020)

\bibitem{deng2019arcface}
Deng, J., Guo, J., Xue, N., Zafeiriou, S.: Arcface: Additive angular margin
  loss for deep face recognition. In: Proceedings of the IEEE/CVF conference on
  computer vision and pattern recognition. pp. 4690--4699 (2019)

\bibitem{deng2019accurate}
Deng, Y., Yang, J., Xu, S., Chen, D., Jia, Y., Tong, X.: Accurate 3d face
  reconstruction with weakly-supervised learning: From single image to image
  set. In: Proceedings of the IEEE/CVF Conference on Computer Vision and
  Pattern Recognition Workshops. pp.~0--0 (2019)

\bibitem{DECA:Siggraph2021}
Feng, Y., Feng, H., Black, M.J., Bolkart, T.: Learning an animatable detailed
  {3D} face model from in-the-wild images. vol.~40 (2021),
  \url{https://doi.org/10.1145/3450626.3459936}

\bibitem{gao2021information}
Gao, G., Huang, H., Fu, C., Li, Z., He, R.: Information bottleneck
  disentanglement for identity swapping. In: Proceedings of the IEEE/CVF
  Conference on Computer Vision and Pattern Recognition. pp. 3404--3413 (2021)

\bibitem{goodfellow2014generative}
Goodfellow, I., Pouget-Abadie, J., Mirza, M., Xu, B., Warde-Farley, D., Ozair,
  S., Courville, A., Bengio, Y.: Generative adversarial nets. Advances in
  neural information processing systems  \textbf{27} (2014)

\bibitem{isola2017image}
Isola, P., Zhu, J.Y., Zhou, T., Efros, A.A.: Image-to-image translation with
  conditional adversarial networks. In: Proceedings of the IEEE conference on
  computer vision and pattern recognition. pp. 1125--1134 (2017)

\bibitem{karras2017progressive}
Karras, T., Aila, T., Laine, S., Lehtinen, J.: Progressive growing of gans for
  improved quality, stability, and variation. arXiv preprint arXiv:1710.10196
  (2017)

\bibitem{karras2019style}
Karras, T., Laine, S., Aila, T.: A style-based generator architecture for
  generative adversarial networks. In: Proceedings of the IEEE/CVF Conference
  on Computer Vision and Pattern Recognition. pp. 4401--4410 (2019)

\bibitem{karras2020analyzing}
Karras, T., Laine, S., Aittala, M., Hellsten, J., Lehtinen, J., Aila, T.:
  Analyzing and improving the image quality of stylegan. In: Proceedings of the
  IEEE/CVF Conference on Computer Vision and Pattern Recognition. pp.
  8110--8119 (2020)

\bibitem{kemelmacher2016transfiguring}
Kemelmacher-Shlizerman, I.: Transfiguring portraits. ACM Transactions on
  Graphics (TOG)  \textbf{35}(4), ~1--8 (2016)

\bibitem{kim2021exploiting}
Kim, H., Choi, Y., Kim, J., Yoo, S., Uh, Y.: Exploiting spatial dimensions of
  latent in gan for real-time image editing. In: Proceedings of the IEEE/CVF
  Conference on Computer Vision and Pattern Recognition. pp. 852--861 (2021)

\bibitem{kim2021smooth}
Kim, J., Lee, J., Zhang, B.T.: Smooth-swap: A simple enhancement for
  face-swapping with smoothness. arXiv preprint arXiv:2112.05907  (2021)

\bibitem{lee2018diverse}
Lee, H.Y., Tseng, H.Y., Huang, J.B., Singh, M., Yang, M.H.: Diverse
  image-to-image translation via disentangled representations. In: Proceedings
  of the European conference on computer vision (ECCV). pp. 35--51 (2018)

\bibitem{li2019faceshifter}
Li, L., Bao, J., Yang, H., Chen, D., Wen, F.: Faceshifter: Towards high
  fidelity and occlusion aware face swapping. arXiv preprint arXiv:1912.13457
  (2019)

\bibitem{liu2019radam}
Liu, L., Jiang, H., He, P., Chen, W., Liu, X., Gao, J., Han, J.: On the
  variance of the adaptive learning rate and beyond. In: Proceedings of the
  Eighth International Conference on Learning Representations (ICLR 2020)
  (April 2020)

\bibitem{mescheder2018training}
Mescheder, L., Geiger, A., Nowozin, S.: Which training methods for gans do
  actually converge? In: International conference on machine learning. pp.
  3481--3490. PMLR (2018)

\bibitem{mosaddegh2014photorealistic}
Mosaddegh, S., Simon, L., Jurie, F.: Photorealistic face de-identification by
  aggregating donors’ face components. In: Asian Conference on Computer
  Vision. pp. 159--174. Springer (2014)

\bibitem{na2019miso}
Na, S., Yoo, S., Choo, J.: Miso: Mutual information loss with stochastic style
  representations for multimodal image-to-image translation. arXiv preprint
  arXiv:1902.03938  (2019)

\bibitem{naruniec2020high}
Naruniec, J., Helminger, L., Schroers, C., Weber, R.M.: High-resolution neural
  face swapping for visual effects. In: Computer Graphics Forum. vol.~39, pp.
  173--184. Wiley Online Library (2020)

\bibitem{park2019semantic}
Park, T., Liu, M.Y., Wang, T.C., Zhu, J.Y.: Semantic image synthesis with
  spatially-adaptive normalization. In: Proceedings of the IEEE/CVF Conference
  on Computer Vision and Pattern Recognition. pp. 2337--2346 (2019)

\bibitem{richardson2021encoding}
Richardson, E., Alaluf, Y., Patashnik, O., Nitzan, Y., Azar, Y., Shapiro, S.,
  Cohen-Or, D.: Encoding in style: a stylegan encoder for image-to-image
  translation. In: Proceedings of the IEEE/CVF Conference on Computer Vision
  and Pattern Recognition. pp. 2287--2296 (2021)

\bibitem{rossler2019faceforensics++}
Rossler, A., Cozzolino, D., Verdoliva, L., Riess, C., Thies, J., Nie{\ss}ner,
  M.: Faceforensics++: Learning to detect manipulated facial images. In:
  Proceedings of the IEEE/CVF International Conference on Computer Vision. pp.
  1--11 (2019)

\bibitem{Ruiz_2018_CVPR_Workshops}
Ruiz, N., Chong, E., Rehg, J.M.: Fine-grained head pose estimation without
  keypoints. In: The IEEE Conference on Computer Vision and Pattern Recognition
  (CVPR) Workshops (June 2018)

\bibitem{sanyal2019learning}
Sanyal, S., Bolkart, T., Feng, H., Black, M.J.: Learning to regress 3d face
  shape and expression from an image without 3d supervision. In: Proceedings of
  the IEEE/CVF Conference on Computer Vision and Pattern Recognition. pp.
  7763--7772 (2019)

\bibitem{tov2021designing}
Tov, O., Alaluf, Y., Nitzan, Y., Patashnik, O., Cohen-Or, D.: Designing an
  encoder for stylegan image manipulation. ACM Transactions on Graphics (TOG)
  \textbf{40}(4),  1--14 (2021)

\bibitem{ulyanov2016instance}
Ulyanov, D., Vedaldi, A., Lempitsky, V.: Instance normalization: The missing
  ingredient for fast stylization. arXiv preprint arXiv:1607.08022  (2016)

\bibitem{wang2021high}
Wang, T., Zhang, Y., Fan, Y., Wang, J., Chen, Q.: High-fidelity gan inversion
  for image attribute editing. arXiv preprint arXiv:2109.06590  (2021)

\bibitem{wang2021hififace}
Wang, Y., Chen, X., Zhu, J., Chu, W., Tai, Y., Wang, C., Li, J., Wu, Y., Huang,
  F., Ji, R.: Hififace: 3d shape and semantic prior guided high fidelity face
  swapping. arXiv preprint arXiv:2106.09965  (2021)

\bibitem{wu2021stylespace}
Wu, Z., Lischinski, D., Shechtman, E.: Stylespace analysis: Disentangled
  controls for stylegan image generation. In: Proceedings of the IEEE/CVF
  Conference on Computer Vision and Pattern Recognition. pp. 12863--12872
  (2021)

\bibitem{xia2022gan}
Xia, W., Zhang, Y., Yang, Y., Xue, J.H., Zhou, B., Yang, M.H.: Gan inversion: A
  survey. IEEE Transactions on Pattern Analysis and Machine Intelligence
  (2022)

\bibitem{yoo2019coloring}
Yoo, S., Bahng, H., Chung, S., Lee, J., Chang, J., Choo, J.: Coloring with
  limited data: Few-shot colorization via memory augmented networks. In:
  Proceedings of the IEEE/CVF Conference on Computer Vision and Pattern
  Recognition. pp. 11283--11292 (2019)

\bibitem{zhang2019lookahead}
Zhang, M., Lucas, J., Ba, J., Hinton, G.E.: Lookahead optimizer: k steps
  forward, 1 step back. Advances in Neural Information Processing Systems
  \textbf{32} (2019)

\bibitem{zhang2018unreasonable}
Zhang, R., Isola, P., Efros, A.A., Shechtman, E., Wang, O.: The unreasonable
  effectiveness of deep features as a perceptual metric. In: Proceedings of the
  IEEE conference on computer vision and pattern recognition. pp. 586--595
  (2018)

\bibitem{zhu2017unpaired}
Zhu, J.Y., Park, T., Isola, P., Efros, A.A.: Unpaired image-to-image
  translation using cycle-consistent adversarial networks. In: Proceedings of
  the IEEE international conference on computer vision. pp. 2223--2232 (2017)

\bibitem{zhu2021one}
Zhu, Y., Li, Q., Wang, J., Xu, C.Z., Sun, Z.: One shot face swapping on
  megapixels. In: Proceedings of the IEEE/CVF Conference on Computer Vision and
  Pattern Recognition. pp. 4834--4844 (2021)

\end{thebibliography}
\end{document}